\documentclass[runningheads]{llncs}

\usepackage{eccv}

\usepackage{eccvabbrv}

\usepackage{graphicx}
\usepackage{wrapfig}
\usepackage{booktabs}
\usepackage{caption}
\usepackage{adjustbox}

\usepackage[accsupp]{axessibility}  

\usepackage{xcolor}
\usepackage{mathtools}

\newif\ifdraft
\draftfalse

\ifdraft
\newcommand{\dcc}[1]{{\color{red}[\textbf{Danny:} #1]}}
\newcommand{\avc}[1]{{\color{purple}[\textbf{Andrey:} #1]}}
\newcommand{\ahc}[1]{{\color{orange}[\textbf{Amir:} #1]}}

\newcommand{\todo}[1]{{\color{blue}[\textbf{TODO:} #1]}}

\newcommand{\dc}[1]{{\color{red}#1}}
\newcommand{\av}[1]{{\color{purple}#1}}

\newcommand{\drop}[1]{}

\else
\newcommand{\dcc}[1]{}
\newcommand{\avc}[1]{}
\newcommand{\cqc}[1]{}
\newcommand{\ahc}[1]{}
\newcommand{\todo}[1]{}
\newcommand{\dc}[1]{{\color{black}#1}}
\newcommand{\av}[1]{{\color{black}#1}}

\fi

\makeatletter
\DeclareRobustCommand\onedot{\futurelet\@let@token\@onedot}
\def\@onedot{\ifx\@let@token.\else.\null\fi\xspace}
\DeclareMathAlphabet\mathbfcal{OMS}{cmsy}{b}{n}

\makeatother

\makeatletter
\def\blfootnote{\xdef\@thefnmark{}\@footnotetext}
\makeatother

\newif\ifwatermark
\watermarktrue
\draftfalse

\usepackage{hyperref}

\usepackage{orcidlink}

\begin{document}

\title{Curved Diffusion: A Generative Model With Optical Geometry Control} 

\titlerunning{Curved Diffusion}

\newif\ifarxiv
\arxivtrue
\newcommand{\condarxiv}[2]{%
    \ifarxiv
        #1
    \else
        #2
    \fi
}

\makeatletter
\newcommand{\printfnsymbol}[1]{%
  \textsuperscript{\@fnsymbol{#1}}%
}
\makeatother

\makeatletter
\renewcommand*{\@fnsymbol}[1]{\ensuremath{\ifcase#1\or *\or \dagger\or \ddagger\or
   \mathsection\or \mathparagraph\or \|\or **\or \dagger\dagger
   \or \ddagger\ddagger \else\@ctrerr\fi}}
\makeatother

\author{Andrey Voynov\inst{1} \and
Amir Hertz\inst{1} \and
Moab Arar\inst{2}\thanks{Performed this work while working at Google.} \and
Shlomi Fruchter\inst{1}\thanks{Equal advising contribution.} \and
Daniel Cohen-Or\inst{2}\printfnsymbol{1}\printfnsymbol{2}}

\authorrunning{A.~Voynov et al.}

\institute{Google \and Tel Aviv University}

\maketitle

\begin{abstract}
State-of-the-art diffusion models can generate highly realistic images based on various conditioning like text, segmentation, and depth. However, an essential aspect often overlooked is the specific camera geometry used during image capture. The influence of different optical systems on the final scene appearance is frequently overlooked. This study introduces a framework that intimately integrates a text-to-image diffusion model with the particular lens geometry used in image rendering. Our method is based on a per-pixel coordinate conditioning method, enabling the control over the rendering geometry. Notably, we demonstrate the manipulation of curvature properties, achieving diverse visual effects, such as fish-eye, panoramic views, and spherical texturing using a single diffusion model.
\end{abstract}

\section{Introduction}

Professional photographers rely on a diverse array of camera lenses, each suited to specific photography needs, from fish-eye lenses for expansive, wide-angle shots to macro lenses for capturing minute details. These lenses, when paired with the same camera kit, yield vastly different images. However, in the realm of text-to-image diffusion models, achieving similar control is more complex. Currently, the most common method involves explicitly including the lens type in the text prompt, which offers only a rudimentary approximation of the nuanced lens-based experience and not feasible to applications that require an exact lens fidelity, such as VR.


\begin{figure}[h!]
    \begin{subfigure}{0.49\textwidth}
        \includegraphics[width=\linewidth]{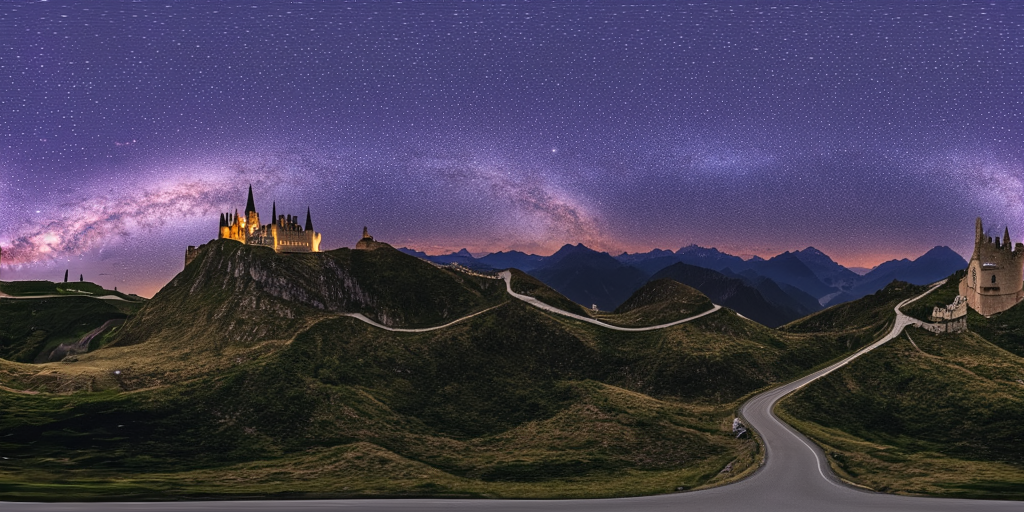}
    \end{subfigure}
    \begin{subfigure}{0.49\textwidth}
        \includegraphics[width=\linewidth]{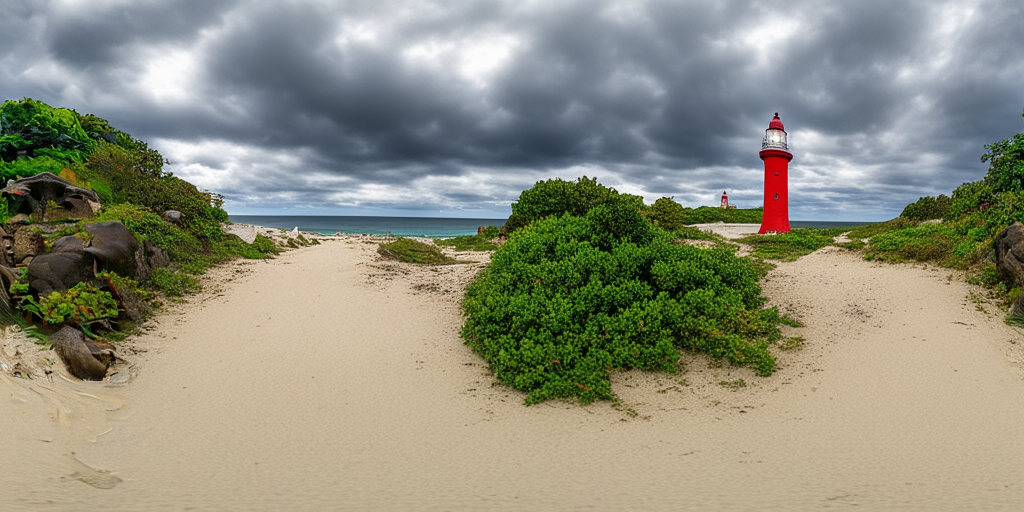}
    \end{subfigure}

    \begin{subfigure}{0.24\textwidth}
        \includegraphics[width=\linewidth]{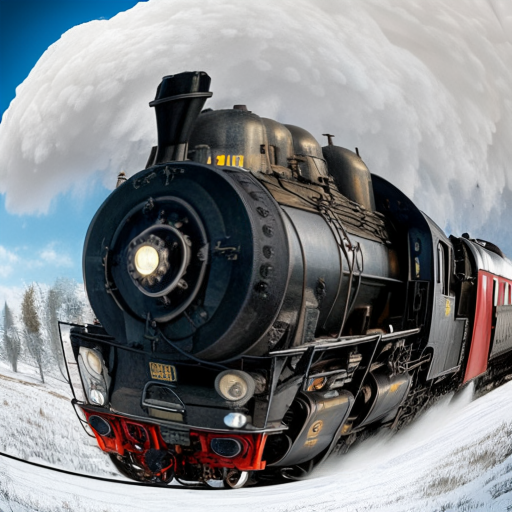}
    \end{subfigure}
    \begin{subfigure}{0.24\textwidth}
        \includegraphics[width=\linewidth]{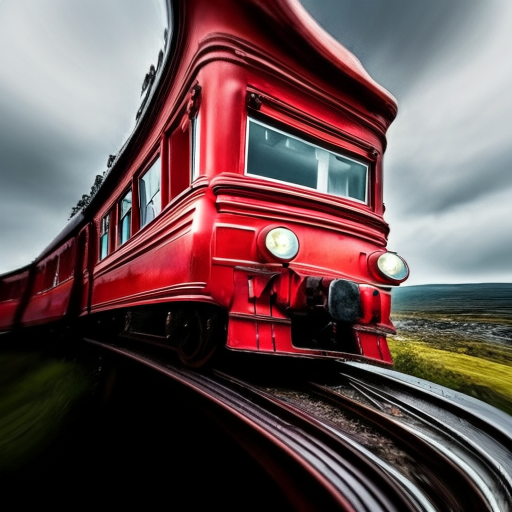}
    \end{subfigure}
    \begin{subfigure}{0.48\textwidth}
        \includegraphics[width=\linewidth]{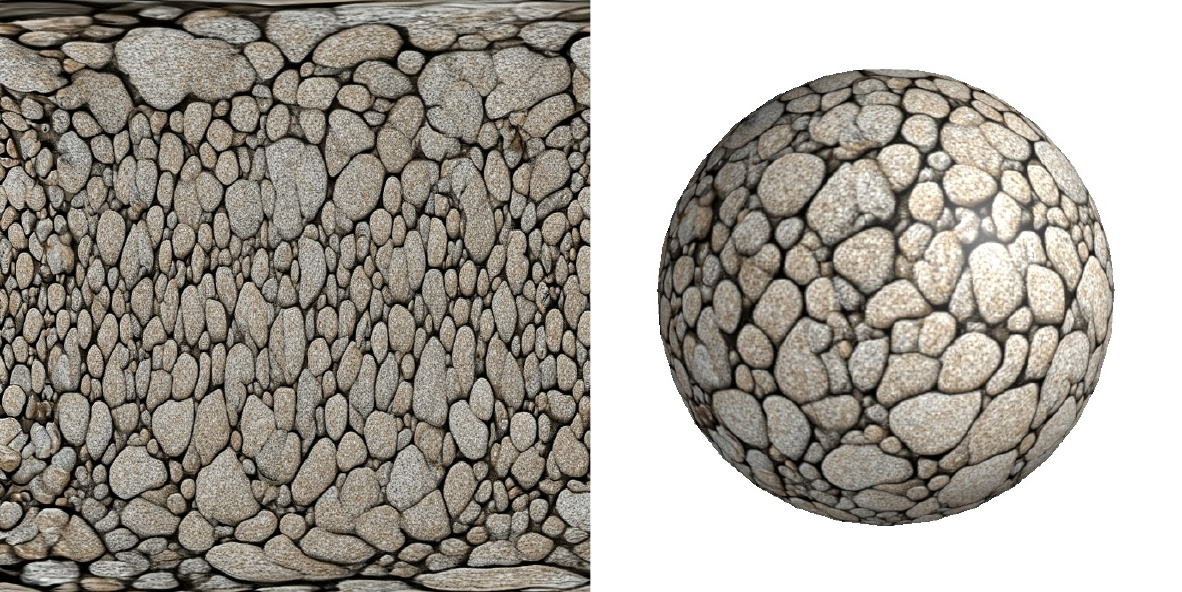}
    \end{subfigure}

    \caption{Images generated with different lens warps. \textit{Top row}: unwarped 3d-sphere stereo panoramas; \textit{bottom row:} fisheye lens, concave lens, sphere texturing.}
    \label{fig:teaser}
    \vspace{-15pt}
\end{figure}

In our work, we address the challenge of incorporating diverse optical geometry controls into a text-to-image diffusion model. \av{Figure \ref{fig:teaser} shows generated images aligned with different geometrical conditions using the method proposed in this work.} We introduce a novel method that trains a text-to-image diffusion model with the ability to condition on local lens geometry, thereby enhancing the model's ability to capture and replicate intricate optical effects.

\begin{figure}[h!]
    \centering
    \includegraphics[width=0.75\textwidth]{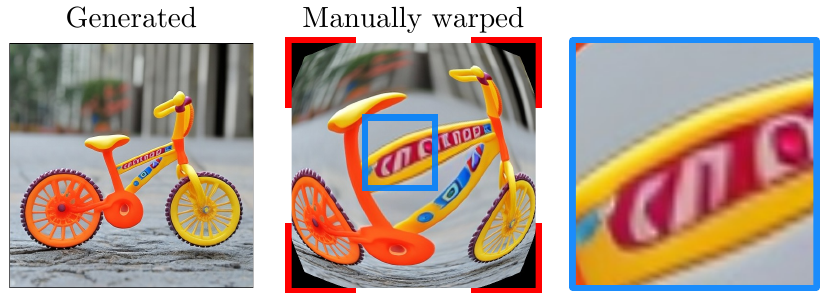}
    \caption{
    Applying a lens geometry transformation in a post-process after the generation of the image produces low-quality image in highly expanded regions. Moreover, the corner regions which are behind the default image canvas are left uncovered.
    }
    \label{fig:naive}
\end{figure}

\av{An alternative approach to reproduce a certain lens effect is to apply the corresponding image deformation over a generated image. In the assumption that a generated image corresponds to some default camera projection with no distortion, we may perform the standard remapping defined by a new lens. This operation can be represented as a composition of a projective image transformation and distortion. Effectively this is a dense image transformation with some of the pixels on the resulted image being uncovered, and some of the regions of the original pixels being upscaled with a resolution loss as illustrated in Figure \ref{fig:naive}. Further more, this approach is limited only to lens geometries that could be represented as a rectangular grid warping, and are not able to generate more complex scenes like a 3d-sphere unwarping.}

\vspace{-18pt}
\vspace{-15pt}

We go beyond the canvas transformations defined by projective camera models, virtually allowing any grid warps. We implement this with a per-pixel coordinate conditioning provided to the diffusion model, which is the spatial positions of a pixel in an unwarped image. Beyond its use to simulate any camera lens, we show that this per-pixel conditioning enables a wide range of applications such as panoramic \av{scene} generation and sphere texturing with a general manifold geometry-aware image generation framework with metric tensor conditioning which is not achievable by warping a rectified image.

\section{Related works}

\subsection{Controllable Generation with Text-to-Image Diffusion Models}

Recent advancements in image generation tasks are built on large-scale diffusion models that are trained on paired text and image data \cite{dalle, imagen, latent_diffusion}. To improve the level of visual control over the generated content there were introduced a variety conditioning approaches by either explicit model conditioning \cite{classifier_free_guidance} or gradient guidance \cite{dhariwal2021diffusion, voynov2023sketch}. A number of work perform conditioning by concatenating the input noised image with an unchangeable conditioning tensor \cite{palette, latent_diffusion}. More sophisticated approaches \cite{controlnet, t2i_adapter} modify a pretrained model architecture to make it conditional to some new modalities. These methods typically fine-tune a text-conditioned model on a set of extracted image features (e.g., depth, segmentation) with the task of reconstructing the original image from a text prompt together with these visual features.

\subsection{Varying Shape Image Generation}

While most image generation approaches assume that the training and target data samples have fixed resolution, a number of works reported numerous advantages by going beyond fixed form. In particular, Chai et al. \cite{anyres} consider the GANs training framework that utilizes the information of original training samples resolutions. They couple each image with the original image resolution information, without rescaling it to some fixed size. This enables generation of images at multiple resolutions and shapes with the same model.

A number of works address similar scope of problems in diffusion models. In particular, \cite{sdxl} train a text-to-image diffusion model with data augmented with crops. They introduce extra conditioning on the crop location on the original image which improves model quality and also allows to modulate the generated image scale. Bar-Tal et al. \cite{multidiffusion} propose an approach to generate large images of resolution much higher than models' default output, with a pretrained text-to-image diffusion model. They create a larger image by assembling smaller tiles, ensuring that the content aligns at the intersections of the tiles. Zhang et al. \cite{zhange2023diffcollage} propose a method to generate large images composed of several smaller, with a graph of overlapping components. They also demonstrate cubemap environment generation with conditioning to segmentation cubemap. Jin et al. \cite{trainingfree_resize} notice that the diffusion tends to produce corrupted images when the generation process runs in resolution different from the training resolution. Their work proposes the self-attention scale adjustment to match the training and inference attention probabilities entropy.

Unlike the above works, where all the synthesized images follow the regular grid geometry, we consider a much wider space of free deformations. We reveal model capabilities based on different families of conditioning per-pixel warps.


\subsection{Generative Texture}
Deep generative priors of vision based models are adapted for text conditioned 3D asset creation.
Recent works show remarkable capabilities of optimizing a neural radiance field (NeRF) \cite{nerf} that matches an input text using supervision from T2I diffusion model \cite{poole2022dreamfusion, metzer2022latent}. Another group of works utilize T2I or CLIP models for optimizing a texture that fit an input 3D shape and text \cite{text2mesh, gao2022get3d, TEXTure, Tang2023mvdiffusion}.
However, these methods require long optimization and a neural renderer that fill the domain gap between distorted texture maps to the distribution of real images. To the best of our knowledge, our method is the first that can be used directly for text conditioned spherical panorama generation and texturing with accurate surface curvature alignment.
\section{Method}

\begin{figure}
    \centering
    \includegraphics[width=\textwidth]{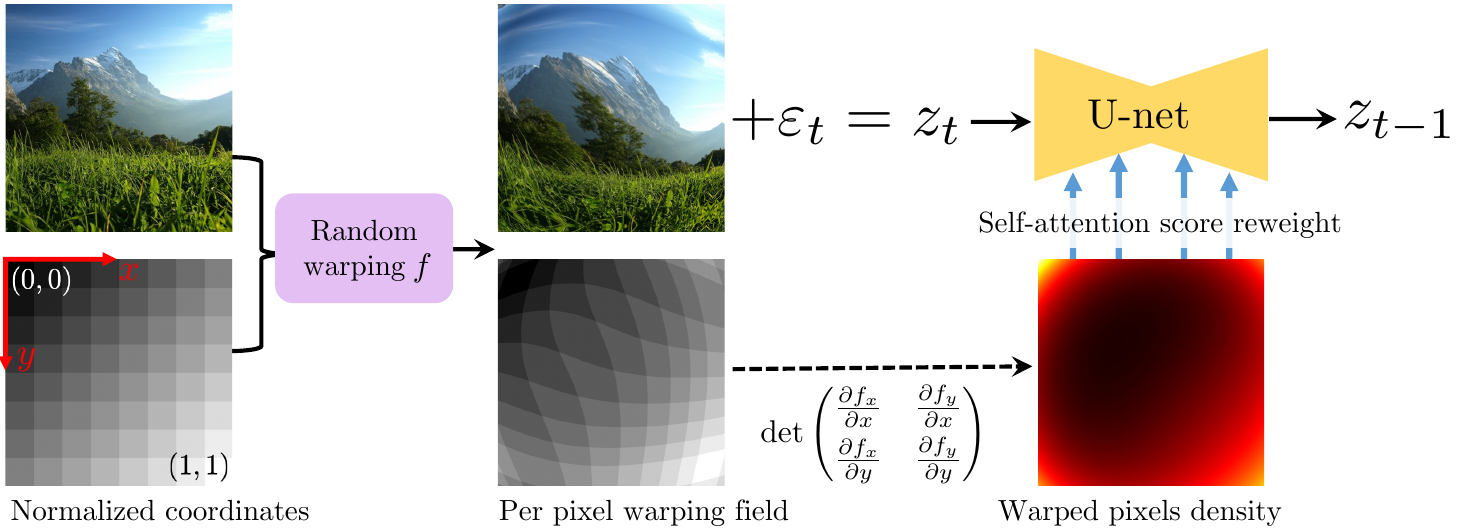}
    \caption{The method training scheme. The training sample is processed with a random distortion, applied over both the image and normalized coordinates grid. Then, the warped image is noised with an additive noise $\varepsilon_t$ correspondent to the denoising step $t$. The denoising U-net model takes the noised image concatenated with the warping field and predicts the denoised image. All the self-attention layers weights are re-weighted according to the original image pixels density.}
    \label{fig:train-scheme}
\end{figure}

Our goal is to enable the generation of images with arbitrary curved lens and camera geometry. Hence, we consider a more general problem, where we aim to allow the model to generate arbitrary warped images. The approach is based on conditioning each of the generated pixels to its location in the standard undistorted camera view.
The input noised image concatenated with the conditioned warping field forms a two-channel array which is used as the input to the diffusion denoising model, as shown in Figure \ref{fig:train-scheme}. This figure also illustrates the denoising U-net that gets a warping field as a spatial condition. As shown in the figure, the U-net also gets a map that re-weights its internal self-attention layers to compensate for sparse regions (we shall elaborate more in the following).

The model is trained by fine-tuning a pre-trained latent diffusion model with training data that we generate by simply warping images with a random warping field as illustrated in Figure \ref{fig:train-scheme}. Below we describe the implementation details for data generation, training and model modifications, and inference strategy.

\subsection{Training Data}



The training data consists of a large textually annotated images dataset, and a per-pixel warping field randomly generated for each of the images.
We assume that the training images are all undistorted, and were generated with the same diagonal projection camera matrix, with the focal center located in the image center.

\begin{wrapfigure}{r}{0.4\textwidth}
\vspace{-20pt}
    \centering
    \includegraphics[width=0.39\textwidth]{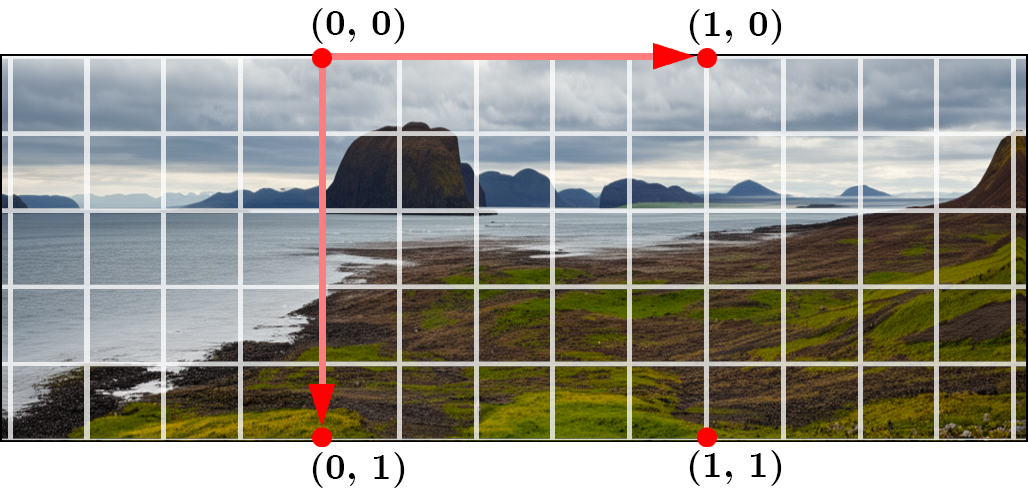}
    \caption{Normalized coordinates in the image. The coordinates unit square is set to be the centered maximum square crop.}
    \label{fig:norm_coords}
    \vspace{-20pt}
\end{wrapfigure}

To unify images of different resolutions or aspect ratios, an image of a size $H \times W$ is accomplished with the uniform coordinates grid with a unit square be the centered maximum square crop (Figure \ref{fig:norm_coords}).
Formally, for a pixel at the location $(i_x, i_y)$ we compute its normalized position as $(x, y) = (\frac{i_x}{\min(H, W)} + s_x, \frac{i_y}{\min(H, W)} + s_y)$. Once $W > H$ the shift terms $s_x, s_y$ are equal $(\frac{W - H}{2}, 0)$, and $(0, \frac{H - W}{2})$ vice versa.

Following the standard lens distortion model \cite{brown1996decentering, opencv}, we sample the camera focal center and the leading distortion coefficients $k_1, k_2, p_1, p_2$ that parameterize it, and apply the resulting distortion to the original image, and perform a random square crop of the distorted image. Simultaneously we apply the same distortion and crop to the normalized coordinates grid of pixel positions on the original image. In our data pipeline, we \textit{first}: randomly warp original data sample; \textit{second}: crop and resize to the models' resolution. As data samples are commonly of a higher resolution than the model resolution ($512 \times 512$) the resulting samples don't have the up-scaling artifacts in the expanded regions. Figure \ref{fig:train-scheme}, \textit{left}, illustrates this coordinates-aware random distortion data preprocessing. Notably, even though on the training stage we perform only the warping described above, the model generalizes to handle different grid transformations which are not representable as a lens distortion.

\subsection{Self-attention Score Reweighting}




\begin{figure*}[t]
    \centering
    \includegraphics[width=\textwidth]{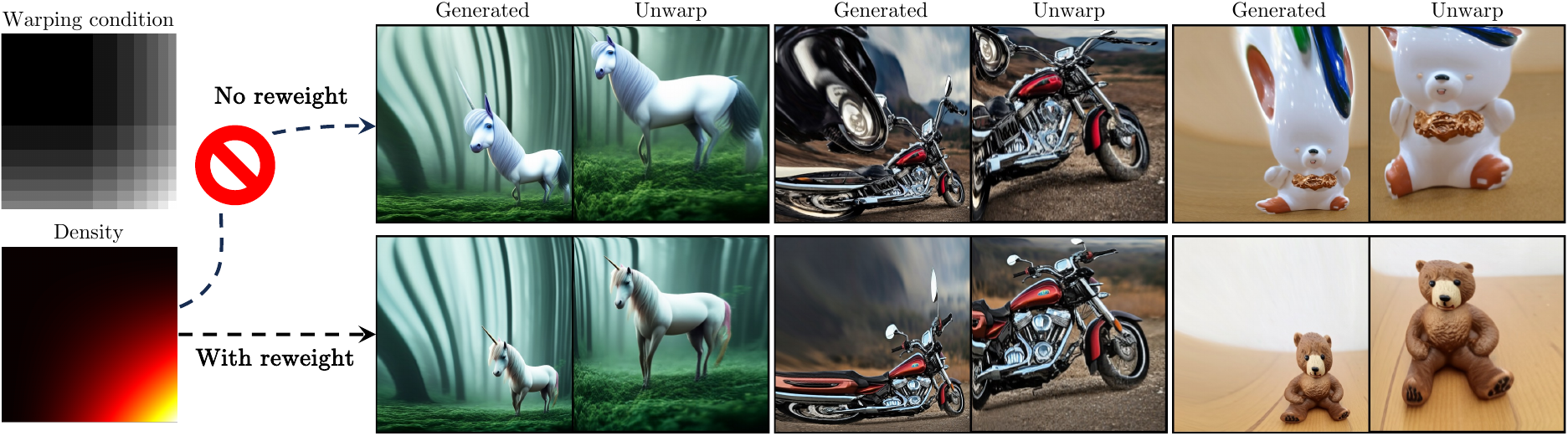}
    \caption{Image generated without (top row), and with (bottom row) self-attention reweighting. For each of the samples we also depict its unwarping. No reweighting induces unproportional object parts (first and third examples),  hallucinations in low-density regions (second example), and overall quality degradation.}
    \label{fig:reweight_ablation}
    \vspace{-10pt}
\end{figure*}

One may notice that warping also changes the density of the actual image content. In highly expanding regions the actual information-per-pixel is lower compared to highly-contracting regions. As internally the diffusion model encapsulates several spatial self-attention layers, this suggests that this should also affect the self-attention mechanism where tokens associated with different image locations share information with each other. Intuitively, a spatial token in the region with a larger density should take higher attention, and vice versa.

To account for the density of a region in the image, we adjust the respective attention given to that region. Let us remind how the self-attention mechanism works in detail. There are three sets of vectors: queries $\left[q_1, \dots, q_N\right]$, keys $\left[k_1, \dots, k_N\right]$, and values $\left[v_1, \dots, v_N\right]$. For the $i$-th token, the self-attention layer output is calculated as
$o_i = \sum\limits_{j=1}^{N} w_{ij} v_j$. The weights quantities $w_{ij}$ are defined as the outputs of the softmax, calculated over the dot products between the query $q_i$ and all the keys. The softmax inputs are also named as scores $s_{il} = \left<q_i, k_l\right>$:

\begin{equation}
w_{ij} = \frac{\exp(s_{ij})}{\sum\limits_{l = 1}^N \exp(s_{il})}
\label{eq:sa_weight}
\end{equation}

Now given a density $d_n$ associated with a spatial token $n$, we proportionally scale the attention into it. If the density is an integer value, the rescale operation is equal to duplicating the token as many times as the density value (the correspondent calculations are provided in supplementary). That is, for higher density we aim to emulate the proportional number of tokens at a particular spatial location. Taking into account that $\alpha \cdot \exp(x) = \exp(x + \ln \alpha)$, we conduct the updated weights formula:

\begin{equation}
    w'_{ij} = 
    \begin{cases} 
    \frac{\exp(s_{ij})}{\exp(s_{in} + \ln d_n) + \sum\limits_{l \neq n} \exp(s_{il})} & \text{if } j \neq n, \\
    \frac{\exp(s_{in} + \ln d_n)}{\exp(s_{in} + \ln d_n) + \sum\limits_{l \neq n} \exp(s_{il})} & \text{if } j = n.
    \end{cases}
    \label{reweight_j}
\end{equation}


thus we conduct that the token reweighting according to its density is equivalent to the scores reweighting in Equation \ref{eq:sa_weight}: $s'_{ij} = s_{ij}$ if $j \neq n$ and $s'_{in} = s_{in} + \ln d_n$. Performing these calculations to all the tokens, we conclude that the updated scores satisfy
\begin{equation}
s'_{ij} = s_{ij} + \ln d_j
\label{eq:sa_shift}
\end{equation}
That is, the tokens scores are shifted with the logarithm of the density associated with each of them.



Considering the original unwarped image, we assume that the density of the content is spatially uniformly distributed. Once an image is warped with the coordinates warping map $f: (x, y) \to (f_x(x, y), f_y(x, y))$ the push-forward density at the point $(f_x(x, y), f_y(x, y))$ is equal to 
\begin{equation}
d_{f(x,y)} = \det \begin{pmatrix}
\frac{\partial f_x}{\partial x} & \frac{\partial f_y}{\partial x} \\[3pt]
\frac{\partial f_x}{\partial y} & \frac{\partial f_y}{\partial y}
\end{pmatrix}
\label{eq:density}
\end{equation}
in all the self-attention layers of the diffusion denoising model, we use this density to update the self-attention scores following Equation \ref{eq:sa_shift}. We use these scores reweighting both on training and inference steps.

\section{Experiments}

\subsection{Training Details}
We conduct all the experiments over a pretrained latent diffusion model \cite{latent_diffusion} with an architecture similar to Stable Diffusion, which consists of multiple convolutional, self-attention, and cross-attention layers. The default model works with images of a resolution  $512\times 512$ downsampled with a latent encoder to $64\times 64$. We generate random warps with the Brown-Conrady distortion model \cite{brown1996decentering, opencv}. The camera matrix is always diagonal and the focal center point is randomly sampled. The nonzero distortion coefficients $k_1, k_2, p_1, p_2$ are sampled independently, with equal probability of positive and negative distortions. The distributions details are provided in the supplementary materials. We perform conditioning to the 2-channel per-pixel warping field by concatenating it with the input noised latent tensor on every step of the denoising U-net, reserving a more in-depth investigation into a more suitable architecture for future research.

For the self-attention re-weight density in Equation \ref{eq:density} we estimate the partial derivatives with the finite differences formula $\frac{\partial f}{\partial x}|_{(i, j)} \approx \frac{f_{i, j+1} - f_{i, j-1}}{2\Delta}$ with $\Delta$ equal the inverse image width, and similarly for $\frac{\partial f}{\partial y}$. When fine-tuning the base diffusion model with extra concatenated conditioning, as for initialization we set the input convolution kernel weights corresponding to the conditioning to be zero. The fine-tuning is performed with the same optimization parameters as for the original model, for $500 \times 10^3$ steps and batch size $256$. for the first $80\%$ of optimization steps, we sample distortion parameters from smaller warps distribution, and for the last $20\%$ of the steps we sample distortion parameters from more aggressive warps distribution (see supplementary for details).

\subsection{Per-Pixel Conditional Generation}
\label{sec:per-pixel}
In this section, we cover different applications that the per-pixel coordinate conditioning reveals. For the ease of comprehension the images generated with different per-pixel conditionings are showed with a uniform rectangular grid deformed accordingly. We also report the unwarped generated image, calculated with the reverse remap. Once the generated image follows the distortion conditioning, the unwarped image should have the un-distorted geometry with straight lines and equally scaled objects in different spatial locations.

\begin{figure}[htb]
    \centering
    \includegraphics[width=\textwidth]{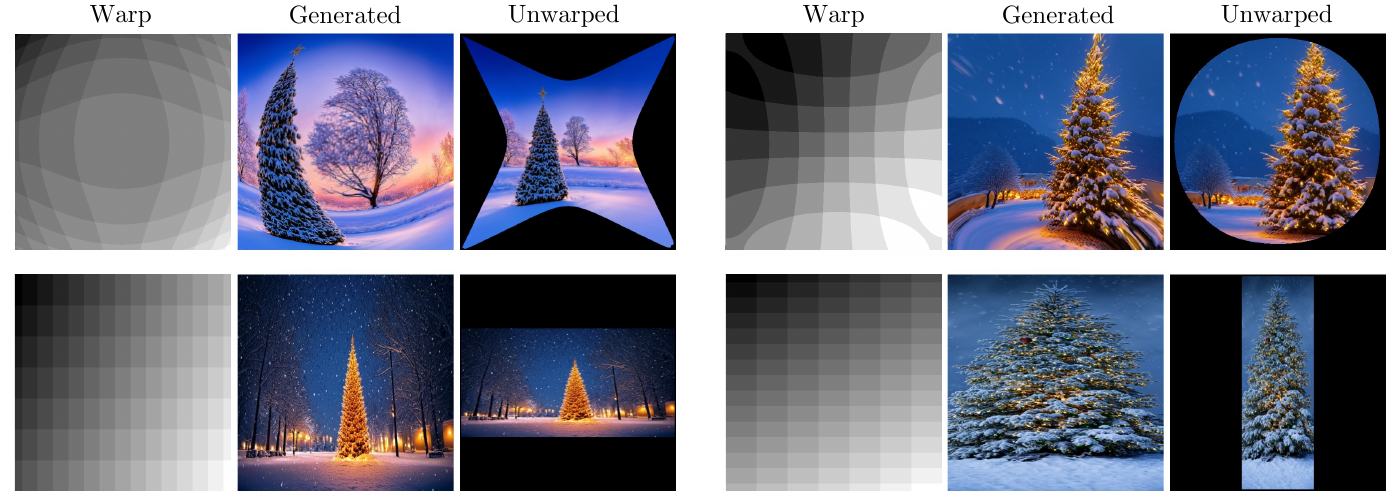}
    \caption{Samples generated with different positional conditionings. \textit{Warp column}: sample grid deformed under the conditional positions. \textit{Generated column}: diffusion model output conditional to the correspondent warped positions and densities, and the prompt \textit{"Professional HD photo of Christmas tree in snow"}. \textit{Unwarped column}: generated image mapped back to the unwarped grid. The generation that precisely follows target geometry should produce fine undistorted images after unwarping.}
    \label{fig:grid}
\end{figure}

\paragraph{Lenses}
We start with qualitative results of the different lenses' geometrical distortions reproduction. The lenses are essentially parameterized with their inner matrix values and the distortion coefficients \cite{brown1996decentering, opencv}. We consider four different lenses: convex and concave with extremely high positive and negative curvature coefficients and extra wide lenses with high reprojection camera focal values. Figure \ref{fig:grid} demonstrates conditional generation with all of these per-pixel conditionings.

\vspace{-15pt}
\vspace{-20pt}

\begin{wrapfigure}{r}{0.5\textwidth}
    \centering
    \vspace{-10pt}
    \includegraphics[width=0.49\textwidth]{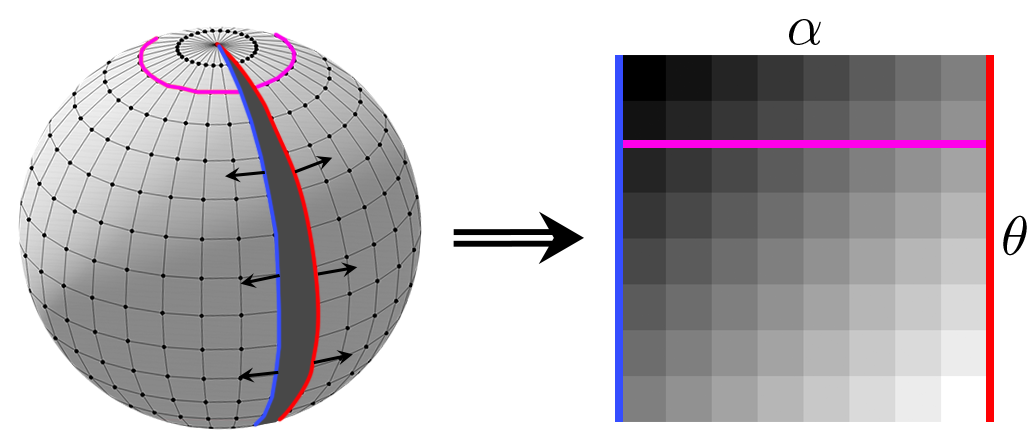}
    \caption{Sphere parametrization with polar coordinates. The horizontal segments $\theta = \mathrm{const}$ lengths are proportional to $\sin{\theta}$.}
    \label{fig:sphere_param}
    \vspace{-20pt}
\end{wrapfigure}

\paragraph{Photospheres}

The per-pixel location conditioning also can be applied over non-planar domains. We illustrate this with the photo-sphere generation and spherical texture generation. We consider the standard unit spherical polar coordinates $(\alpha, \theta)$ with $\alpha \in [0, 2 \pi)$, as shown in Figure \ref{fig:sphere_param}. This sphere parametrization naturally unfolds the sphere to the rectangle $[0, 2\pi] \times [0, \pi]$ with a map $\phi$. For the mapping $\phi$ the partial derivatives satisfy $\|\frac{\partial \phi}{\partial \theta}\| = 1$ and $\|\frac{\partial \phi}{\partial \alpha}\| = \sin^{-1}\theta_0$. That is alongside the grid axis $\phi$ scales tangent vectors with coefficients $1$ and $\sin^{-1}\theta_0$.

Now, given an image of a size $H \times W$, we texture the sphere by mapping each of its pixels at position $(h, w)$ to the unfolded sphere point $(\alpha(w), \theta(h)) = (\frac{w}{W} \cdot 2\pi, \frac{h}{H} \cdot \pi)$, and then applying the inverse $\phi^{-1}$ which is defined almost everywhere. To get well-covering sphere texturing, we aim to find the per-pixel conditioning $c(h, w)$ so that the defined mappings sequence from the $H\times W$ grid to the sphere has unit scaling along the sphere grid lines. This can be done by setting the per-pixel coordinate conditioning at pixel $(h, w)$ correspondent to sphere point $(\alpha, \theta) = (\alpha(w), \theta(h))$ to be equal $(\alpha \cdot \sin(\theta), \theta)$. The sphere volume element in the polar coordinate satisfies $ds^2 = d\theta^2 + \sin^2\theta \cdot d\alpha^2$ resulting the self-attention scale be equal $\ln |\sin\theta(h)|$ at pixel $(h, w)$. We guarantee the generated texture border alignment for $\alpha = 0$ and $\alpha = 2\pi$ by overwriting a small fraction of right vertical border pixels with left border pixels on every denoising step.

Notably, this approach guarantees the correct texture scaling among a pair of directions for each of the points. Although this is enough to obtain plausible results for the waste of applications, in Section \ref{sec:metric} below we discuss an alternative universal solution of this problem.


\begin{figure}[htb]
    \vspace{-10pt}
    \centering
    \includegraphics[width=\textwidth]{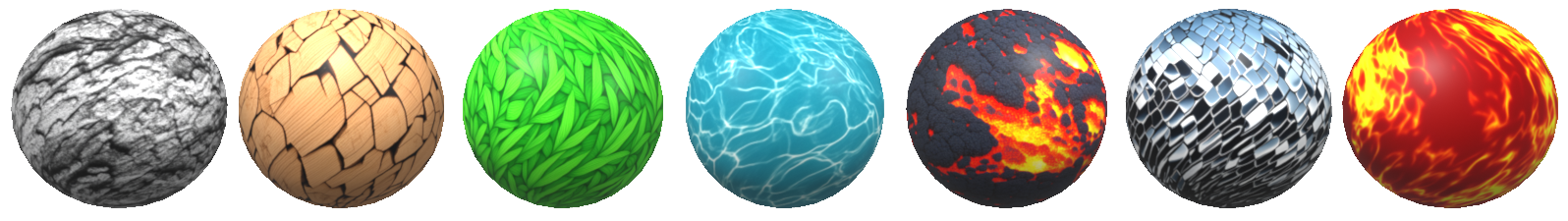}
    \caption{Variety of textures generated with the sphere unfolding.}
    \label{fig:sphere-textures}
    \vspace{-10pt}
\end{figure}

\begin{figure}
  \centering
  \begin{minipage}{0.495\linewidth}
    \includegraphics[width=\linewidth]{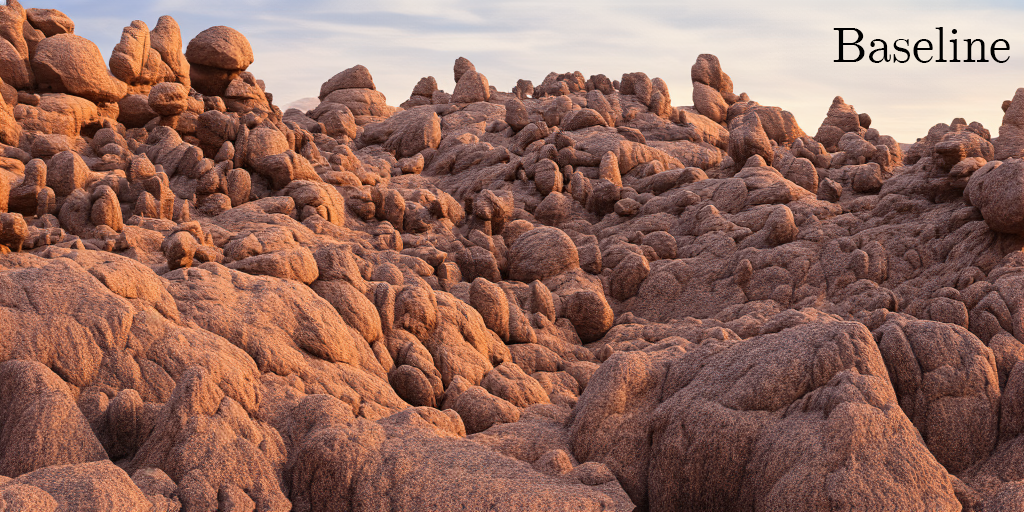}
  \end{minipage}
  \hfill
  \begin{minipage}{0.495\linewidth}
    \includegraphics[width=\linewidth]{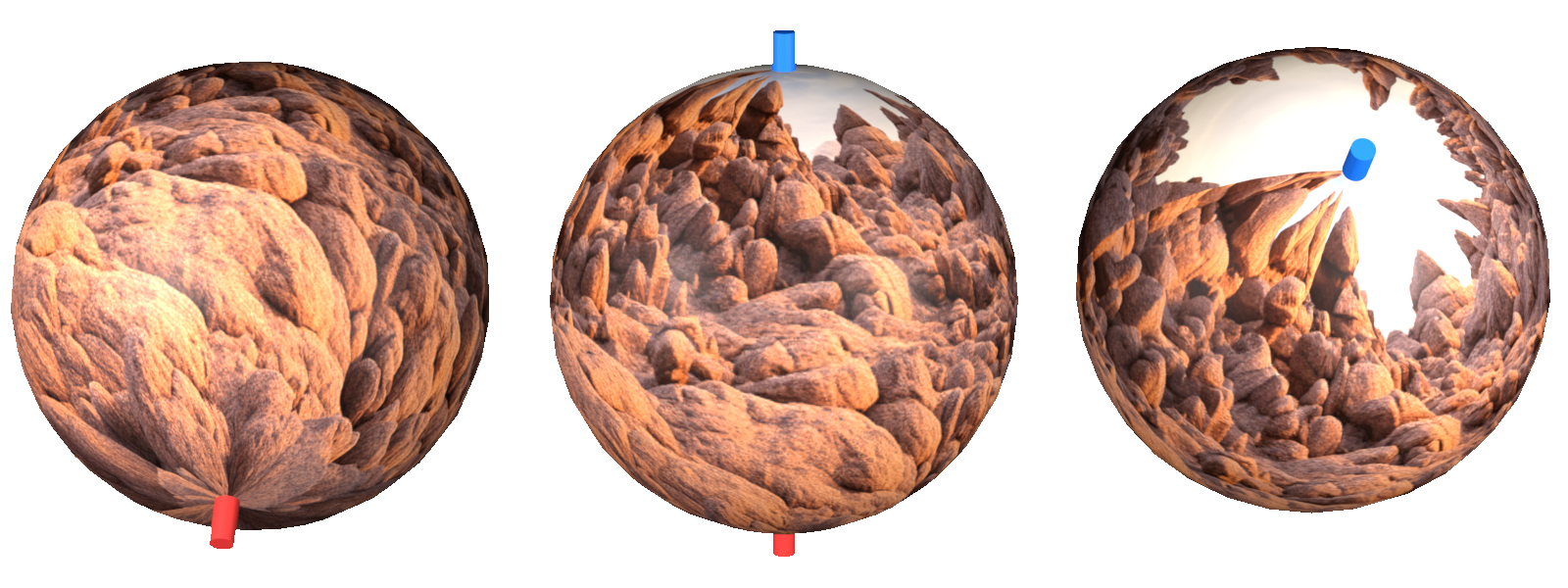}
  \end{minipage}
  
  \begin{minipage}{0.495\linewidth}
    \includegraphics[width=\linewidth]{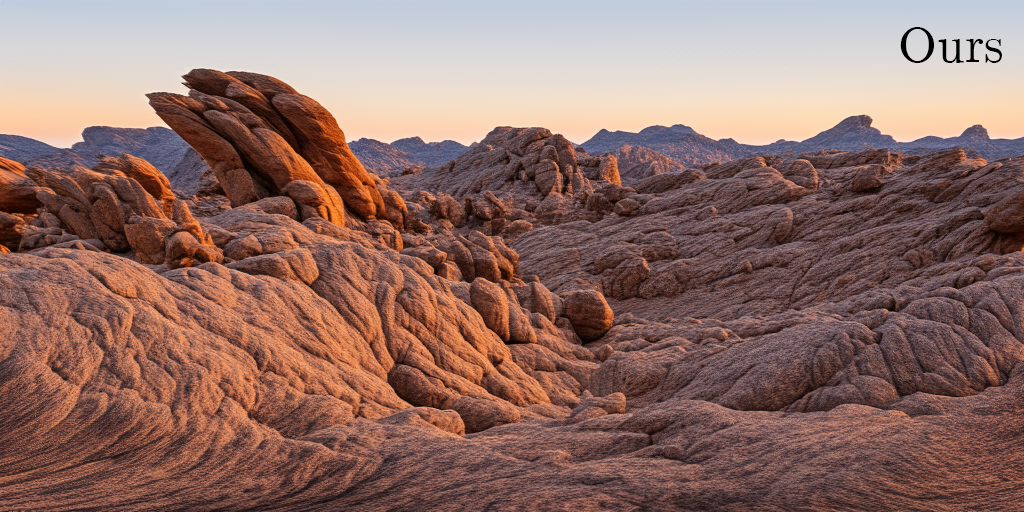} 
  \end{minipage}
  \hfill
  \begin{minipage}{0.495\linewidth}
    \includegraphics[width=\linewidth]{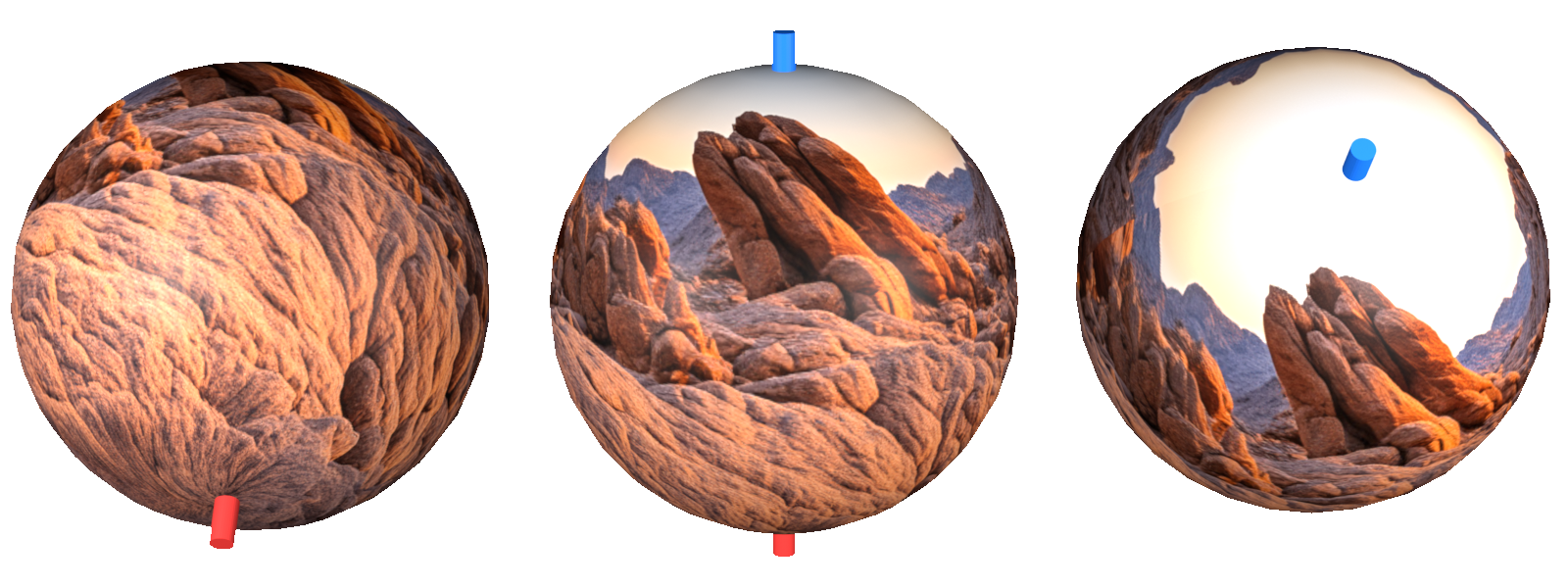} 
  \end{minipage}
    \caption{Spherical panorama generation without (\textit{top row}) and with (\textit{bottom row}) the sphere-aware positional conditioning. The image generated with baseline has unnatural sky proportion and texture-absorbing artifacts on the poles. Prompt: \textit{"rocks landscape"}.}
    \label{fig:sphere_sbs}
    \vspace{-10pt}
\end{figure}

In practice, we take $H = 512, W = 1024$. Figure \ref{fig:sphere-textures} shows multiple sphere textures generated following the described process. The very same way one can generate spherical panorama unfolds. Figure \ref{fig:sphere_sbs} shows the comparison of the proposed sphere texturing approach with the straight-forward image generation with regular diffusion, and then warping generated image onto the sphere with polar coordinates. Notably, our method demonstrates much more realistic image parts proportions and doesn't produce artifacts on the poles.

\paragraph{Role of the Reweight}

Now let us illustrate the role of the self-attention reweighting, introduced in Equation \ref{eq:sa_shift}. Figure \ref{fig:reweight_ablation} presents a batch of images generated with and without the self-attention reweighting. Here we consider the coordinates conditioning with the image being squeezed into the right-bottom corner, causing the top-left corner to have low information density. Notably, once the self-attention reweighting is disabled, the generation process pays extra attention to the low-density region, conducting unpropotional head (first image), artifacts at the low-density region (second image). and overall quality degradation.

\subsection{Further Experiments}

\begin{figure}[h]

    \centering
    \begin{subfigure}{0.48\textwidth}
    \includegraphics[width=\linewidth]{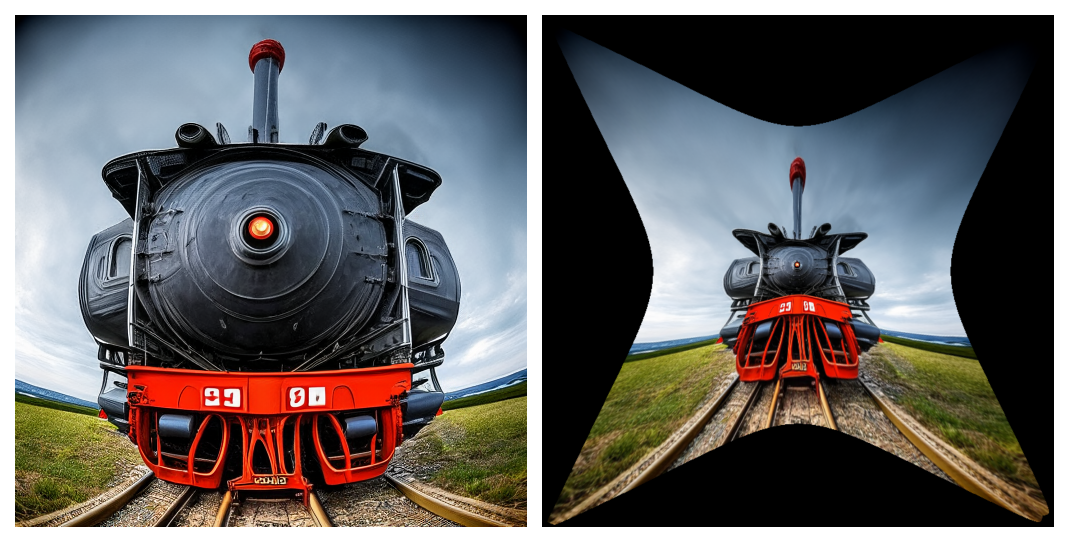}
    \caption{No per-pixel conditioning, prompt: \textit{locomotive, HD \underline{\textbf{fisheye}} photo}}
    \end{subfigure}
    \begin{subfigure}{0.48\textwidth}
    \includegraphics[width=\linewidth]{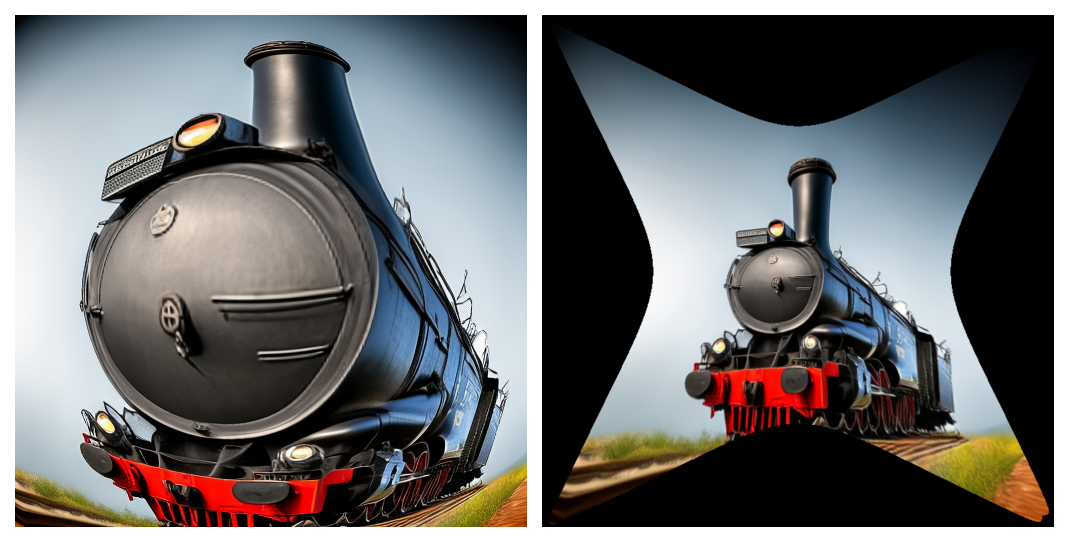}
    \caption{Lens-aware per-pixel conditioning, prompt: \textit{locomotive, HD photo}}
    \end{subfigure}

    \caption{\textit{Left}: image generated with explicit lens description and no proper per-pixel coordinates conditionings; \textit{right}: image generated with the proposed per-pixel coordinates conditionings aligned with the target fisheye lens. The second column represents the undistorted image.}
    \label{fig:text_baseline}
    \vspace{-15pt}
\end{figure}

One may attempt to reproduce the lens-conditioning effect with explicit textual conditioning. Assuming that we have a target fisheye lens parameters, we consider the image generation with the explicit "fisheye" word provided in the prompt. Even though this may look like a fisheye photo, the image would be still miss-aligned with a target lens geometry making it useful for tasks that require precise geometry correctness. Figure \ref{fig:text_baseline} illustrates the two images: in the first row the image is generated with "fisheye" conditioning added to the text, while in the second row, we use the proposed per-pixel coordinates conditioning. Notably, the fisheye effect assured by text only is not aligned with the actual camera geometry. Thus, the undistorted image in the first row still does not look rectified.

Remarkably, while the proposed per-pixel coordinates model is fine-tuned with the initialization from a base diffusion model, it does not compromise the base model quality. Namely, the FID \cite{fid} computed with respect to MS-COCO dataset \cite{mscoco} with the standard $[0, 1] \times [0, 1]$ uniform grid conditioning remains the same as for the base model and equal $17.5$. Further evaluation metrics are presented in Section \ref{sec:quantitative}.

\section{Metric Conditioning}
\label{sec:metric}
As discussed in Section \ref{sec:per-pixel} we still cannot fully accurately generate an image that follows an arbitrary surface geometry. The training data supposes that the surface can be isometrically covered with a plane region. In particular, a sphere is locally non-isometric to a plane, thus it is not possible to define a plane region area with the planar metrics, that can cover a sphere or portion of it.

Thus, here we consider a more general conditioning framework. First, we need to provide a brief overview of the differential geometry concepts we will use. A smooth $d-$dimensional surface in a space (or, more general, a smooth manifold) can be equipped with a distance function defined by the metric tensor $g_{ij}$. This tensor can be represented as $d \times d$ symmetric positive definite matrix which defines the scalar product in each of the surface points tangent planes. While for the standard Euclidean metric in $\mathbb{R}^d$ this tensor is represented by the unit matrix in each of the points, for more complex surfaces it may vary between points. This tensor fully defines the internal surface geometry as well as its curvature (called Gaussian curvature). Once a function $f$ maps a 2-dimensional manifold $\mathcal{M}_1$ with some coordinates $(u_1, u_2)$ to some other manifold $\mathcal{M}_2$ with coordinates $(v_1, v_2)$ and a metrics $g_{ij}$ defined on it, this map induces the pullback metrics $f^*g$ with the rule $(f^*g)_{kl} = g_{ij}\frac{\partial v_i}{u_k}\frac{v_j}{u_l}$ with the summation over all possible $i, j$. \dc{To generate an image that uniformly fits a surface, we need} to find an image and its map to the surface so that in local coordinates it appears casual and unwarped. For a deeper presentation of manifold differential geometry, we refer to \cite{lee2012smooth}.

To make the diffusion generation process flexible for different surfaces metrics, instead of the coordinates conditioning, we introduce the metrics conditioning. As for the training data generation, we use the same warping augmentations, but instead of the deformed normalized coordinates conditioning, for each of the pixels, we condition it to the pullback metrics. Namely, the warping map defines for each of the warped image pixels $(x, y)$ its original normalized coordinates $F(x, y) = (u(x, y), v(x, y))$. Assuming the Euclidean metric $g$ in these normalized coordinates space, the metric $g'$ of the warped image is defined as the pullback $F^*g$. In coordinates this can be written as $g_{11} = (\frac{\partial u}{\partial x})^2 + (\frac{\partial u}{\partial y})^2,\ g_{22} = (\frac{\partial v}{\partial x})^2 + (\frac{\partial v}{\partial y})^2$, and $g_{12} = g_{21} = \frac{\partial u}{\partial x} \frac{\partial u}{\partial y} + \frac{\partial v}{\partial x} \frac{\partial v}{\partial y}$. The partial derivatives are computed with the finite differences. This way we compute the per-pixel metric tensor of the warped image and use it as the new conditioning for the model training. To provide the model the sense of the origin of the coordinates, we also provide as extra conditioning the distance to the image center in the unwarped coordinates $\sqrt{(u(x, y) - 0.5)^2 + (v(x, y) - 0.5)^2}$.

These two conditionings are concatenated and passed to the model just the same way as for the positional conditionings. We train the new model following the same protocol as described above.

\begin{figure}[htb]
    \centering
    
    \adjustbox{valign=c}{
        \includegraphics[width=0.48\textwidth]{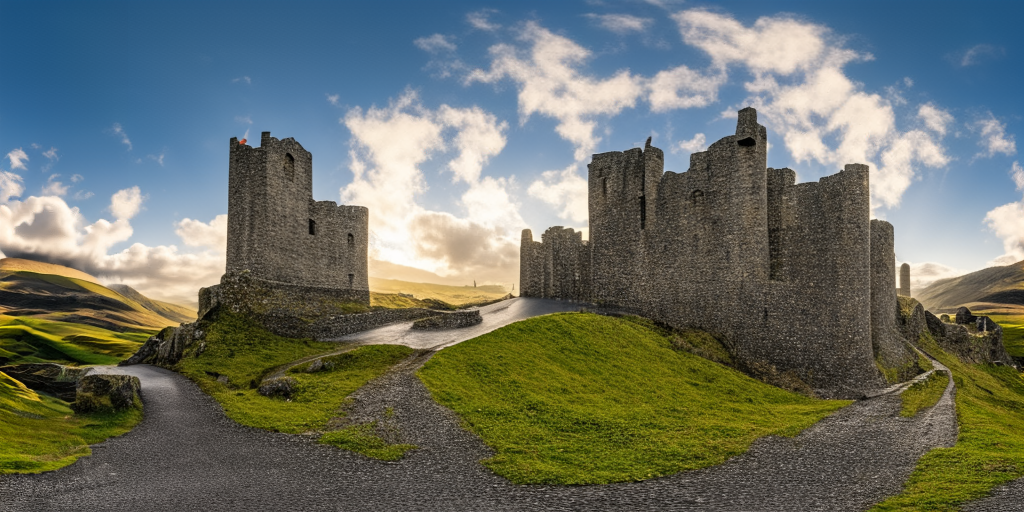}
    }
    \adjustbox{valign=c}{
        \includegraphics[width=0.45\textwidth]{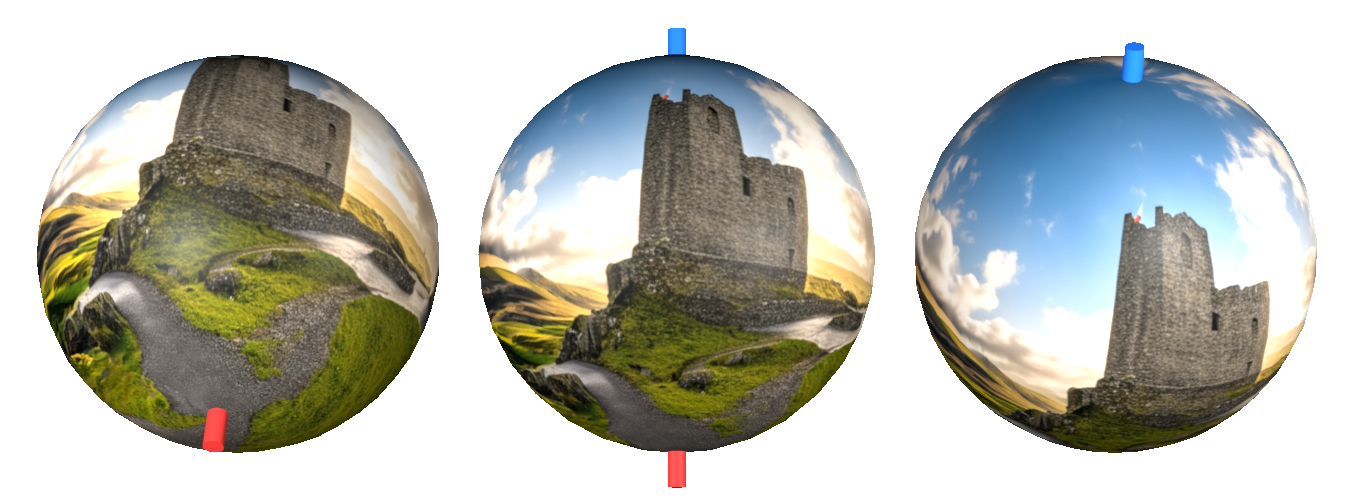}
    }
    \caption{Spherical map generated with the metrics conditioning in polar coordinates. Prompt: \textit{"Castle view, road approaching it, Scotland. Sun shining through clouds"}.}
    \label{fig:sphere-metric}
    \vspace{-20pt}
\end{figure}

\begin{figure}
    \centering
    \includegraphics[width=0.97\textwidth]{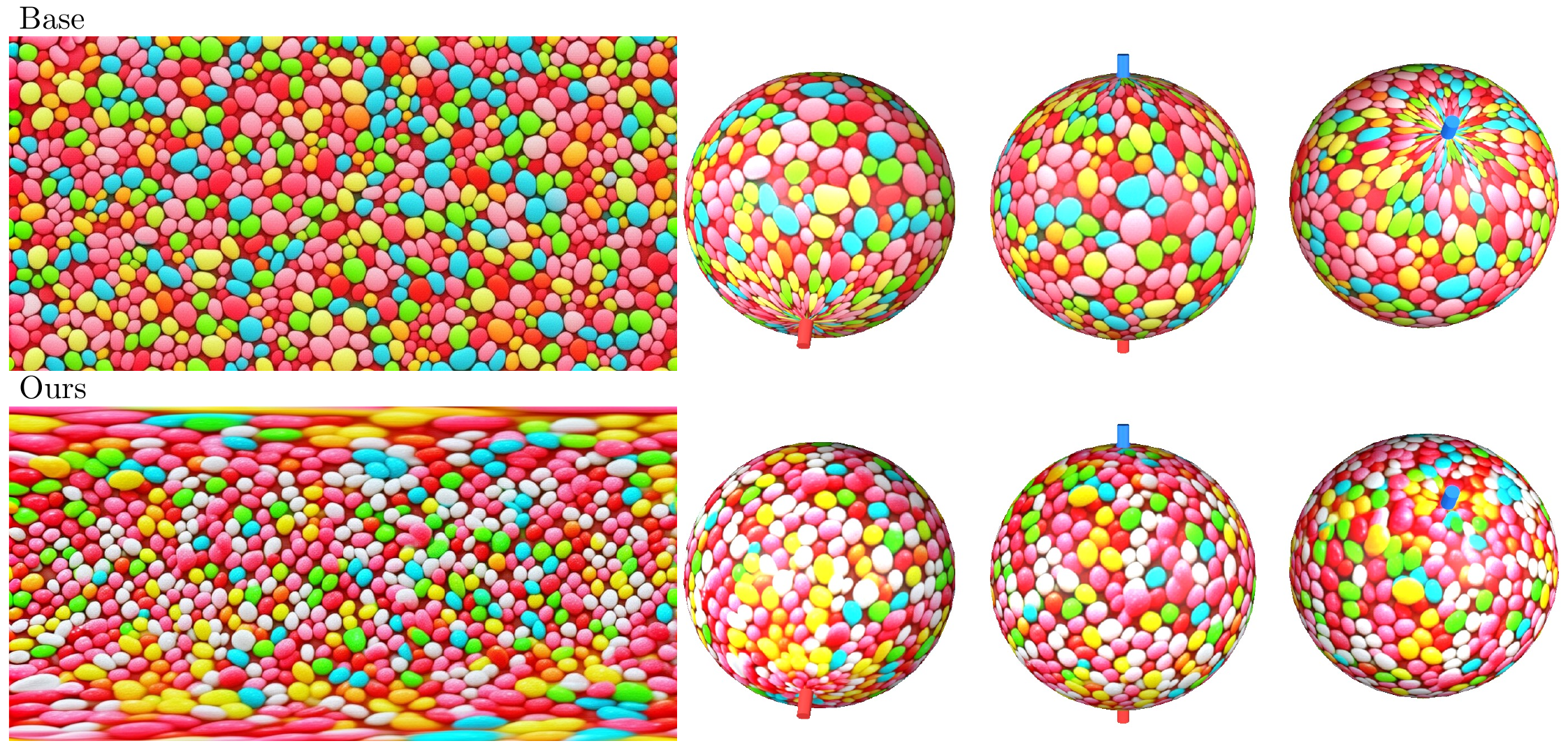}
    \caption{Comparison of the proposed method with metric tensor conditioning and the base text-to-image model sphere texturing. Notably, the images generated with the proposed approach don't have artifacts on the poles and provide uniform sphere coverage.}
    \label{fig:cmp_supp}
\end{figure}

As for the usage example, we describe the inference for the unit sphere panorama generation. Same as in Section \ref{sec:per-pixel} we parameterize the sphere with the polar coordinates $(\alpha, \theta)$ as shown in Figure \ref{fig:sphere_param}. The sphere metrics tensor in these coordinates is written as $g_{ij} = \left(\begin{matrix}
1 & 0 \\
0 & \sin^2\theta \end{matrix}\right)$.
As for self-attention reweight, we also need to compute the sphere volume density in polar coordinates. The volume density element can be calculated as $\sqrt{|\det{g}}|$ which in our case is equal $|\sin\theta|$. As for distance to center calculation, we set the origin to be located at the image center with the polar coordinate $(\pi, \frac{\pi}{2})$. Using the spherical law of cosine, for a point with the polar coordinates $(\alpha, \theta)$ the distance to origin can be computed as $\arccos(|\cos(\alpha - \pi)~\cos(\frac{\pi}{2} - \theta)|)$.
Once all the required conditionings are evaluated, we perform the metrics-conditioned diffusion inference for the spherical panorama generation. One of the examples is presented in Figure \ref{fig:sphere-metric} and Figure \ref{fig:cmp_supp}. This panorama perfectly fits the sphere metrics with no deformations in any direction at any point. Further examples are presented in the supplementary.

\section{Quantitative evaluation}
\label{sec:quantitative}

\subsection{Distortion Fidelity}

\begin{figure}[h]
    \centering
    \vspace{-12pt}
    \includegraphics[width=0.9\textwidth]{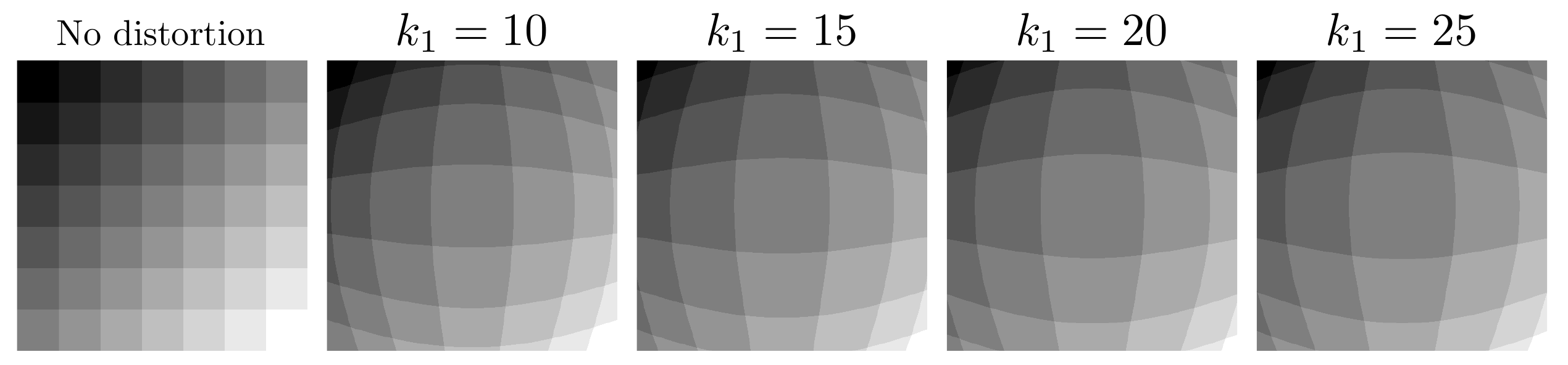}
    \caption{Illustration of distortions used for fidelity study.}
    \label{fig:distortions}
    \vspace{-20pt}
\end{figure}

\begin{figure}[h]
    \centering
    \begin{subfigure}{0.47\textwidth}
        \includegraphics[width=\columnwidth]{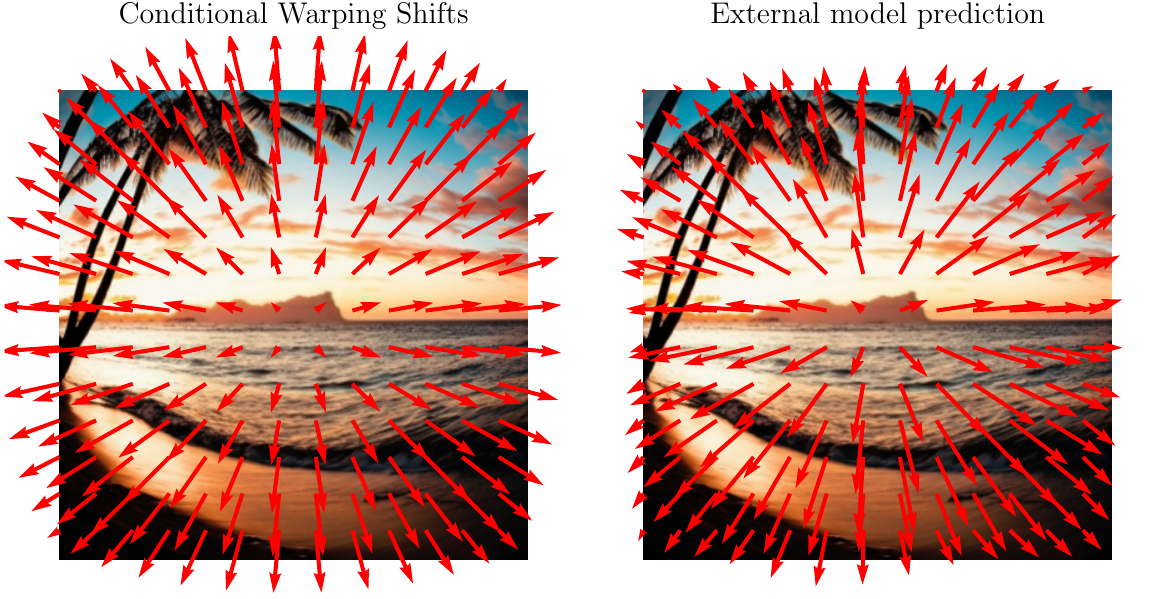}
    \end{subfigure}
    \hfill
    \begin{subfigure}{0.47\textwidth}
        \includegraphics[width=\columnwidth]{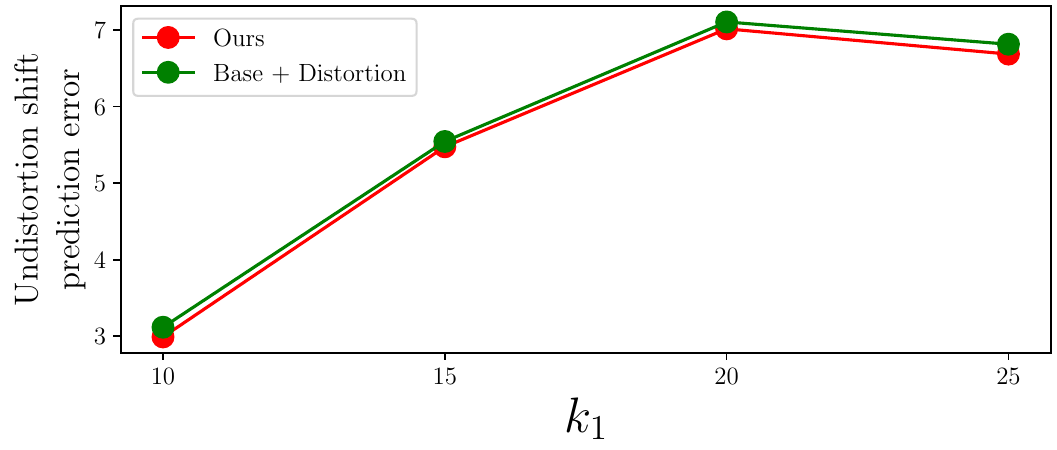}
    \end{subfigure}
    \caption{\textit{Left:} image undistortion shifts calculated based on the conditioning, and predicted by an external pretrained mode. \textit{Right}: error of the external model undistortion prediction (in pixels) for the per-pixel conditional model, and for the undistorted samples of the base text-to-image model.}
    \label{fig:unwarp}
    \vspace{-20pt}
\end{figure}

To quantify the distortion fidelity, we incorporate a pretrained model \cite{undistort} that predicts an undistortion displacement field of a distorted image. Our model generates a $512 \times 512$ image with per-pixel normalized coordinates conditionings $u(x, y), v(x, y)$ for a pixel in location $(x, y)$. Scaling back from normalized coordinates to pixel coordinates, the displacement map that undistorts the generated image is calculated as $(512 \cdot u(x, y) - x, 512 \cdot v(x, y) - y)$. Ideally, the external unwarping model should predict exactly this field, given the generated image with per-pixel locations conditioning, in the assumption of the perfect positional conditioning fidelity, and error-prune undistortion prediction. Both undistortion shifts induced by the conditioning, and predicted by the external model are illustrated in Figure \ref{fig:unwarp} (left). To measure the coordinates conditioning fidelity, we evaluate the averaged $l_2$-norm of the differences between the unwarping models' predictions and the displacement used in conditioning.
We generate images with 100 random prompts, and 8 images generated for each of the prompts. As for the spatial conditionings we take the sequence of 4 fish-eye distortions with the leading curvature coefficients $k_1$ equal $10, 15, 20, 25$ and the rest coefficients set to zero. These distortions are illustrated in Figure \ref{fig:distortions}.

Figure \ref{fig:unwarp} (right) illustrates the averaged norm between the pixels undistortion shifts predicted by the external model, and the shifts calculated based on the conditioning. As the unwarping model may bring its own error in the undistortion estimation, we also report the differences for the images generated with the base text-to-image model, and then implicitly distorted. As the base model produces undistorted images, this difference reflects the external unwarping model error. Notably, our model demonstrates distortion fidelity as good as the manually distorted images which highlights the accuracy in alignment with the per-pixel location conditioning.

\subsection{Human Evaluation}
\label{sec:human_eval}
As for human evaluation, we perform labeling of sphere textures generated with two models. First, the base text-to-image model which was trained with the standard diffusion loss and no per-pixel conditioning. Second, the fine-tuned model conditional to the metrics, as described in Section \ref{sec:metric}. As for texture generation, we use resolution $1024 \times 512$ and both models use the same random seed.

Each time we render a pair of rotatable spheres. A user have to choose: "Which of the spheres is textured better, with less deformations". Optionally, one may skip a particular pair. It is unknown, which model was used to texture the left and the right spheres and they are randomly swapped. We perform labeling of 200 samples generated with 50 different texture prompts, 4 samples per prompt. Each sample is labeled at least once. In average, the proposed metrics-conditioned model was picked $58\%$ times, the base model was picked $12\%$ times, and in the rest $30\%$, a user decided to skip the task. This indicates significantly higher sphere-geometry aware generation quality of the proposed method compared to the baseline.

\section{Conclusions}
Contemporary diffusion models have excelled in generating highly realistic images through various conditioning inputs, yet the influence of diverse optical systems on final scene appearance has often been overlooked. Our work addresses this oversight by integrating lens geometries crucial in image rendering.

By introducing a per-pixel coordinate \av{and metrics} conditioning approach, we have empowered the model to manipulate rendering geometry, granting control over curvature properties. Additionally, our exploration into attention layers led to the presentation of a re-weighting technique that adjusts attention based on the warped density of the image.

The method we presented requires to have a metric representation that is compatible with the geometry of the simulated lens, and it is still limited by the extent of the extrapolation with introducing repetitions. \av{We also leave out of scope the depth of field control.} In the future, we would like to extend our method to overcome these limitations, possibly by exploring more advanced with stronger means for conditioning.

Much like specialized lenses capturing distinct scenes, each contributing to unique visual outcomes, our approach expands the potential for diverse image generation. Enabling the use of any rendering lens opens the door for unexplored applications, encouraging enhanced creativity and control in image synthesis.

\newpage
{\large \textbf{Acknowledgements}}
\\
\\We thank Chu Qinghao, Yael Vinker, Yael Pritch, and Yonatan Shafir for their valuable inputs that helped improve this work. We also express our gratitude to the anonymous reviewers who provided valuable feedback that helped us improve the paper.

%
%
\bibliographystyle{splncs04}
\bibliography{main}

\newpage
{\LARGE{\centering{\textbf{Supplementary}}}}
\setcounter{section}{0} 
\renewcommand{\thesection}{\Alph{section}}

\section{Further Experiments, Examples, and Ablation}

\subsection{Attention Reweighting Analysis}

\begin{figure}[h]
    \centering
    \begin{subfigure}{0.45\textwidth}
        \includegraphics[width=\linewidth]{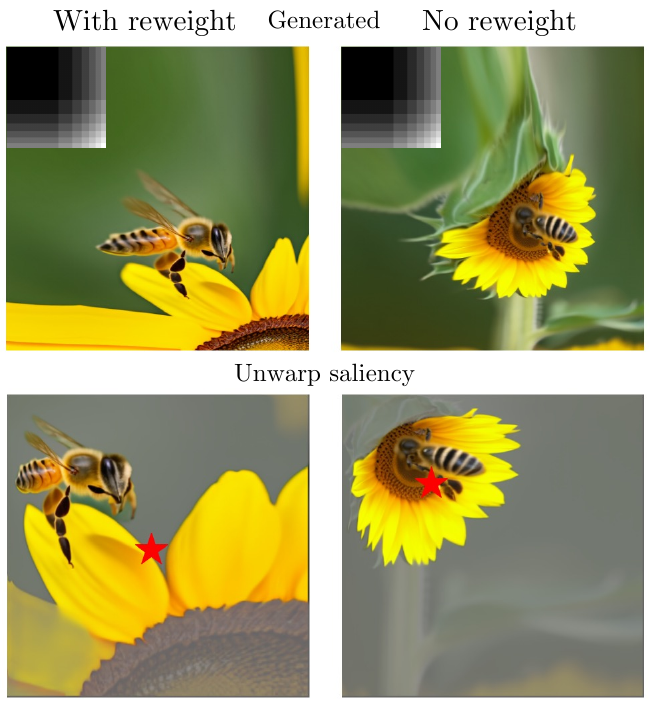}
        \caption{\textit{Top row}: images generated with and without self-attention reweighting; \textit{second row}: saliency object segmentation on the unwarped image, the star indicates the saliency center of mass.}
        \label{fig:saliency_sample}
    \end{subfigure}
    \hfill
    \centering
    \begin{subfigure}{0.45\textwidth}
        \includegraphics[width=\linewidth]{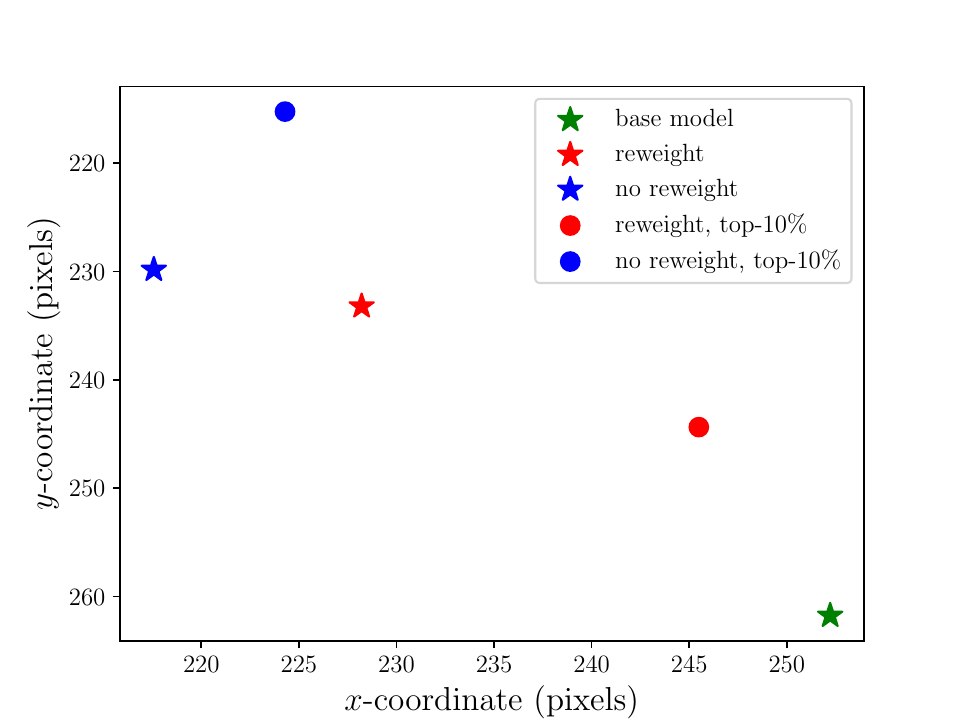}
        \vspace{10pt}
        \caption{Averaged saliency map center location for various generated images. We also depict the saliency centers for the reweighted and non-reweighted generation where the output images are the most different from each other.}
        \label{fig:saliency_scatter}
    \end{subfigure}
    \caption{Analysis of the self-attention reweighting.}
    \condarxiv{}{\vspace{-20pt}}
\end{figure}

To quantify the contribution of the reweighted self-attention introduced in Section 3.2, we perform the following experiment. As for the per-pixel location conditioning, we consider the corner-squeezing deformation, the same as in Figure 5 in the main text. Now we generate two sets of images: one with the model that reweigths the self-attention with respect to the density, and with one that doesn't. For the generation, we use 100 automatically generated prompts, for each of the prompts we sample 8 images. Both models share the random seed. Then we unwrap the generated images with respect to the per-pixel warping conditioning. It is expected that the distribution of objects located on the unwarped image would be the same as the distribution on an image generated with the base text-to-image model, non-conditional to spatial locations. To quantify this alignment, we apply a pretrained saliency object segmentation model \cite{U2net} over the unwarped generated images, as illustrated in Figure \ref{fig:saliency_sample}. We calculate the average saliency map center over the unwarped generated images with reweighting, without reweighting, and for the images generated with the base model.

The corresponding points are depicted in Figure \ref{fig:saliency_scatter}. Notably, the reweighting assures the saliency centers to be better aligned with the expected centers' location. While the original positional conditioning has high distortion in the top-left image region, the model with no reweighting tends to fill it more aggressively, inducing high saliency in the top-left corner of the unwarped image.  Moreover, once we consider the most different images (in per-pixel $l_1$-distance) from the reweighted and non-reweighted generation, one can see that the correspondent alignment and misalignment are even higher. This suggests that higher reweighting impact induces better-generated image fidelity to the positional conditioning.

\subsection{Conditioning Study}

\begin{figure}[h]
    \vspace{-10pt}
    \centering
    \includegraphics[width=\columnwidth]{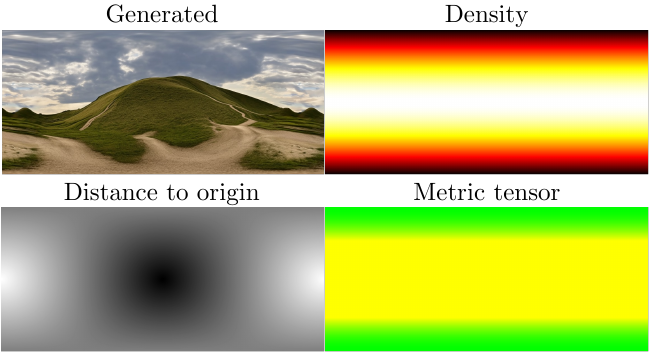}
    \caption{Metrics-conditioned model inputs visualization for the sphere polar coordinates. \textit{Top row}: generated image and the self-attention re-weight density; \textit{bottom row}: spherical distance to the origin, and the spherical metric tensor in polar coordinates, used as conditionings. As for metric visualization, the metric tensor components $g_{11}, g_{22}$ and $g_{12}$ (which is equal to 0) are mapped to RGB.}
    \label{fig:cond_samples}
    \condarxiv{\vspace{-15pt}}{}
\end{figure}

Figure \ref{fig:cond_samples} illustrates the condition inputs for the metric-conditioned model: the density that guides the self-attention reweighting, distance to the origin, calculated as the distance on the sphere, and the spherical metric tensor coefficients in polar coordinates. Notably, all the top row pixels have the same distance to the origin as all of them map to the same point on the sphere.

\begin{figure}
    \condarxiv{\vspace{-10pt}}{\vspace{-10pt}}
    \centering
    \includegraphics[width=\textwidth]{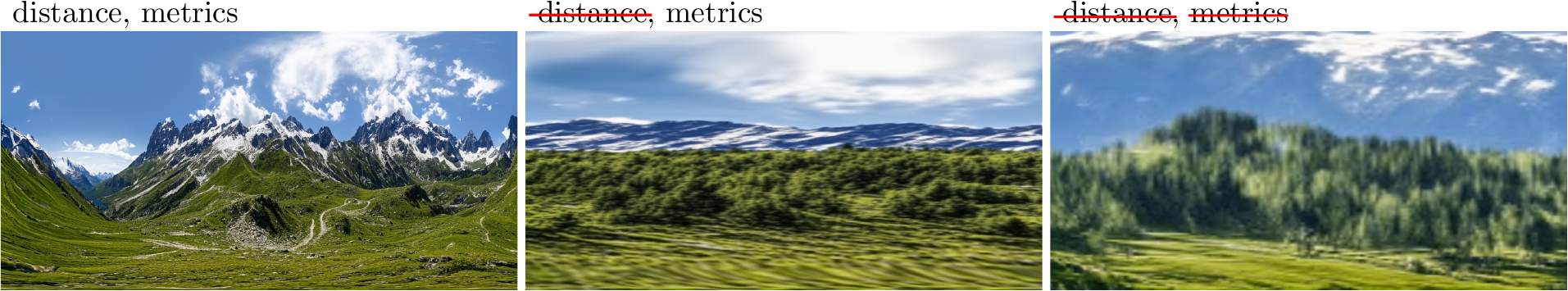}
    \caption{Study of different conditionings roles for the model conditioned on the distance from the origin and the metric tensor. \textit{Left}: model is properly conditioned to both, for sphere polar coordinates; \textit{middle}: model is conditioned to true metrics, and the distance is set to zero; \textit{right}: metrics is set to trivial identical, distance is zero.}
    \label{fig:cond_ablation_on_off}
    \condarxiv{\vspace{-25pt}}{\vspace{-25pt}}
\end{figure}

Figure \ref{fig:cond_ablation_on_off} reveals a qualitative analysis of the metric-conditional model components.
The left image illustrates the generation with the sphere polar coordinates metrics and distance. In middle image generation, we perform generation with zeroed distance. This induces remarkable quality degradation and extreme zoom-in picture, while the metric conditioning still induces varying stretching at different pixels. On the right, we depict the image generated with zeroed distance to center conditioning and trivial diagonal metric. Now the zoom is still extreme and no spherical distortion takes place.

\subsection{Vanishing Point Analysis}
\begin{figure}[h]
    \vspace{-10pt}
    \centering
    \includegraphics[width=\textwidth]{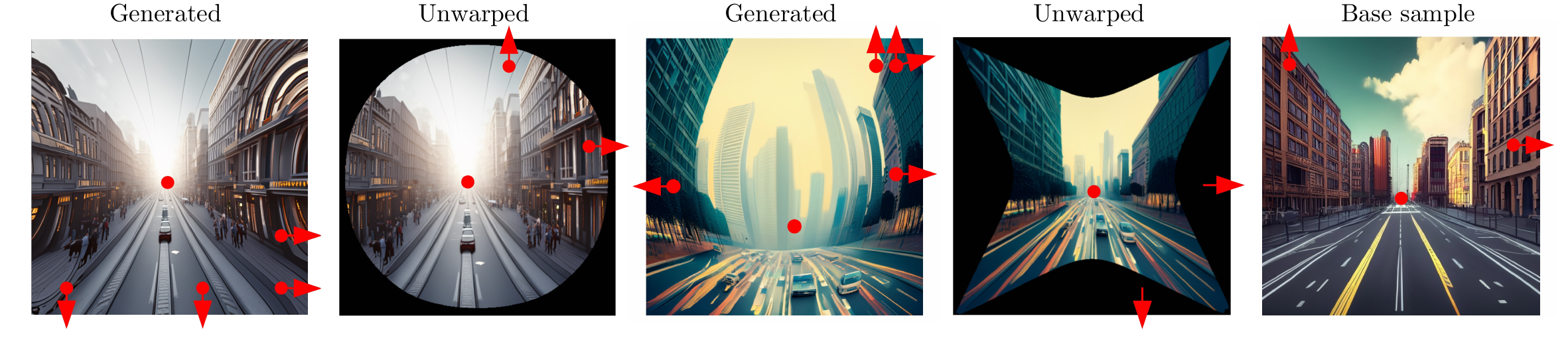}
    \caption{Vanishing points estimated for the images generated with high curvature (marked "Generated"), unwarped generated images (marked "Unwarped"), and the base model sample, as a reference. Red points depicts the estimated vanishing point, the red arrows show the directions to the vanishing points out of the image canvas; Unwarped generated images has three clear vanishing points, similar to baseline generation.}
    \label{fig:vp}
    \vspace{-25pt}
\end{figure}

To examine the fidelity of images generated using a target lens geometry, we perform vanishing point estimation on the unwarped images. Ideally, for straightforward prompts such as 'street view, 3D rendering,' an unwarped image should exhibit clear vanishing points, similar to those in the base text-to-image generation. For this test, we incorporate the vanishing point estimation model from \cite{lin2022vpd}. Figure \ref{fig:vp} illustrates the vanishing point estimation for both the unwarped generated samples and the original generated images, along with a sample from the base model. Once unwarping is performed, the generated images demonstrate distinct vanishing points, thereby ensuring the target lens fidelity.

\subsection{Further quantitative experiments}
We conduct a number of auxiliary quantitative experiments. To have a broader comprehension of the metric conditioning advantage, we perform a user study similar to the one in Section \condarxiv{\ref{sec:human_eval}}{6.2 (main text)} for positional-conditioned spherical textures generation. For positional-conditioning vs baseline labeling, the positional-conditioned generation is picked $38\%$ times, the base model is picked $18\%$ times, and the question is skipped in $34\%$ times. Compared to metric conditioning, position conditioning is picked $28\%$ times, metric conditioning is picked $40\%$ times, and the question is skipped $32\%$ times. This reassures the advantage of position-conditioning over the baseline approach, and the advantage of metric conditioning for non-euclidean curvature warp. 

We also compute COCO-FID score for images, generated with high distortion, same as for the first example in Figure \condarxiv{\ref{fig:grid}}{6 (main text)}. These highly curved images demonstrate FID equal $56.9$; same images after the unwarping demonstrate FID equal $19.8$, and images generated with the base model with the unwrap mask put over it have FID equal $18.6$. One should note that the unwarped samples are degraded by resampling artifacts, which also partially degrades metrics as FID is known to be sensitive to it. Close scores ensure that the method generates images with accurate geometry without compromising quality.

\subsection{Further Examples and Failure Cases}

\begin{figure}[h]
    \vspace{-15pt}
    \centering
    \includegraphics[width=\textwidth]{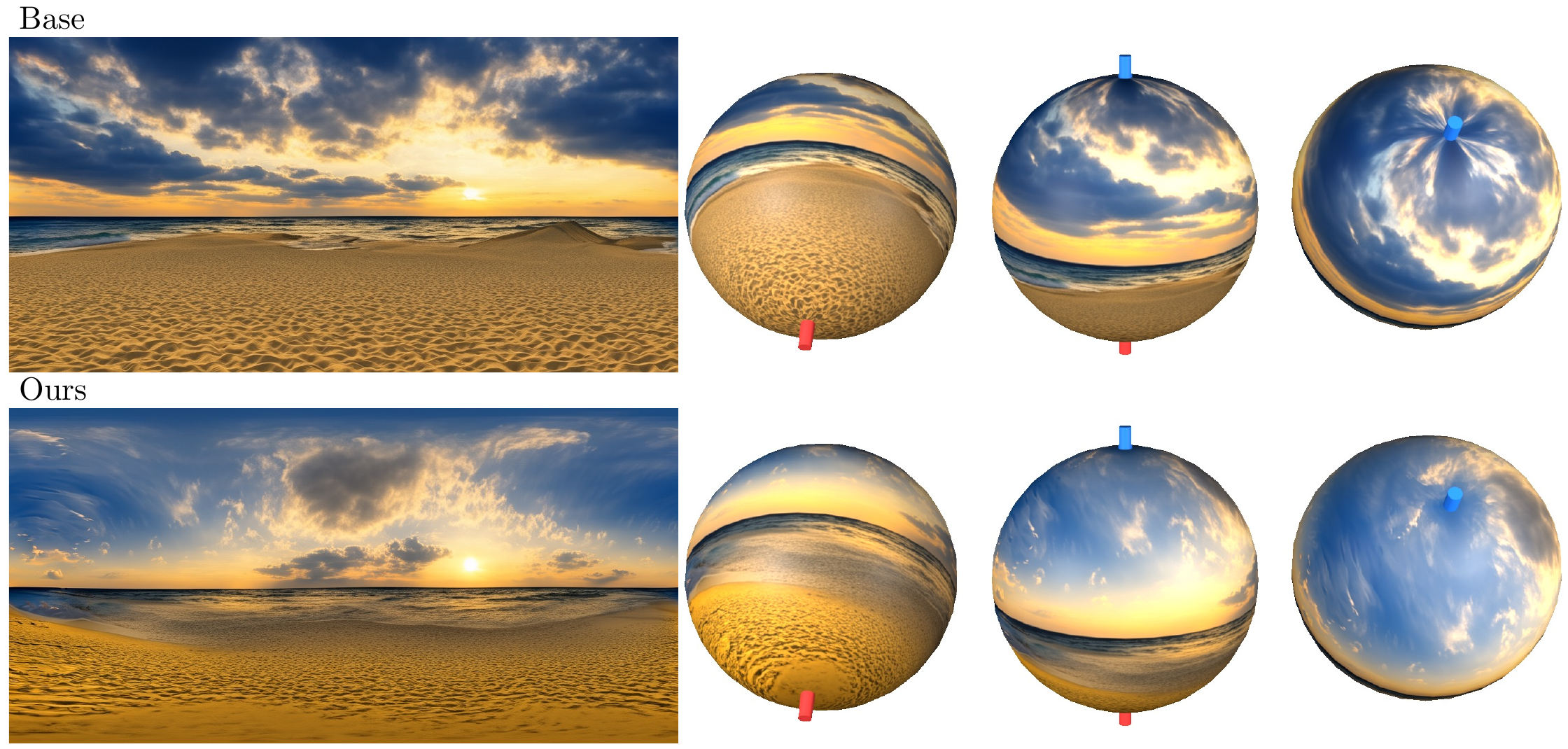}
    \caption{Comparison of the proposed method with metric tensor conditioning and the base text-to-image model sphere texturing. One may note the artifacts on the poles that the baseline produces while our approach provides a uniform sphere coverage that carefully follows curvature at each of the points.}
    \vspace{-15pt}
    \label{fig:cmp_supp_2}
\end{figure}

\begin{figure}[h]
    \centering
    \adjustbox{valign=c}{
    \includegraphics[width=0.48\textwidth]{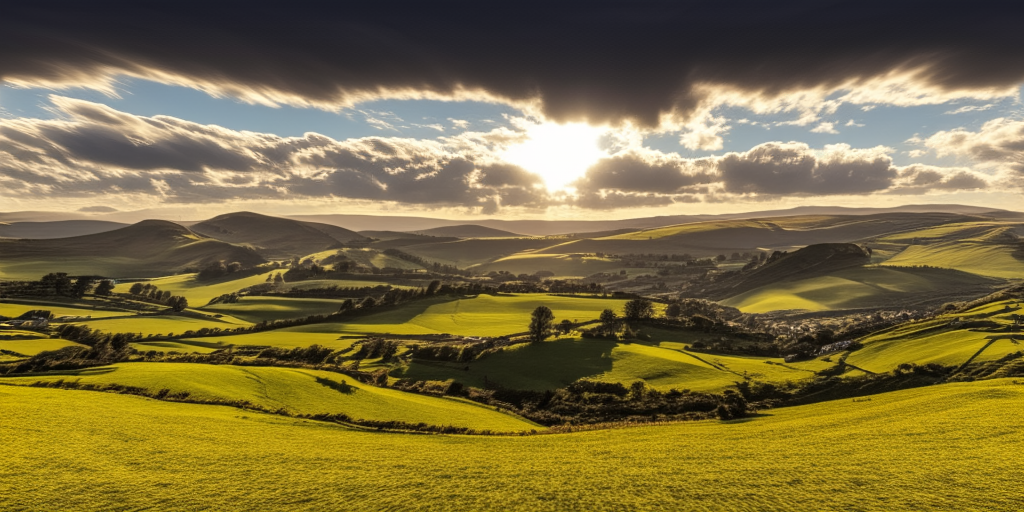}
    }
    \adjustbox{valign=c}{
    \includegraphics[width=0.47\textwidth]{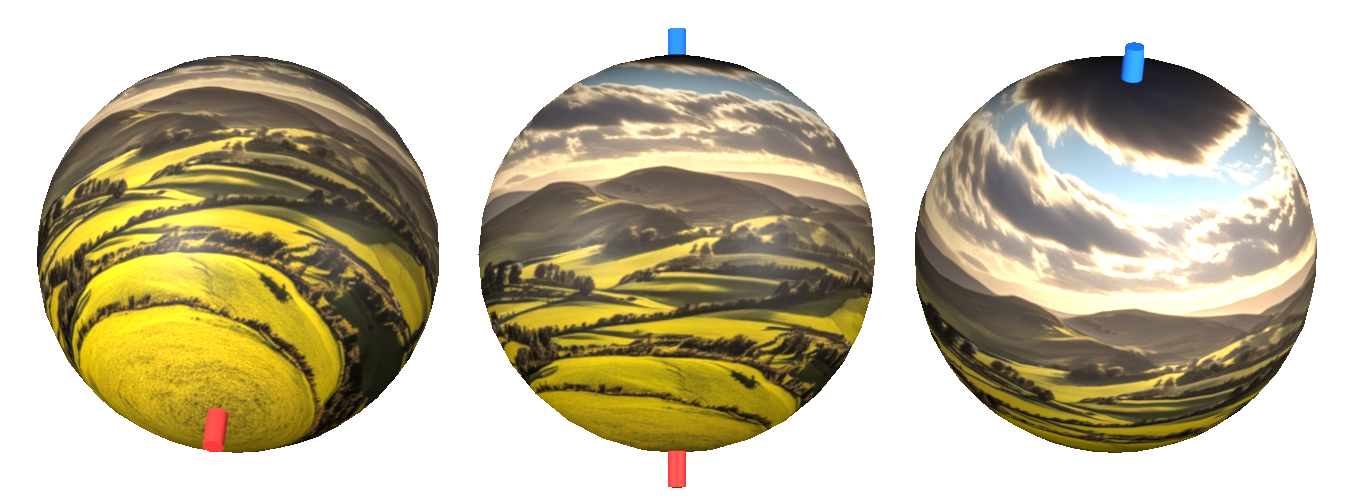}
    }
    \caption{Spherical panoramas generated with the sphere unfolding. Prompt: \textit{"Rolling hills in England, clouds and sun in the sky"}.}
    \label{fig:paonram}
\end{figure}

Figure \ref{fig:cmp_supp_2} shows a further comparison of the proposed method with the baseline model. Figure \ref{fig:paonram} shows another spherical panorama generation example.

\begin{figure}[h!]
    \centering
    \includegraphics[width=\columnwidth]{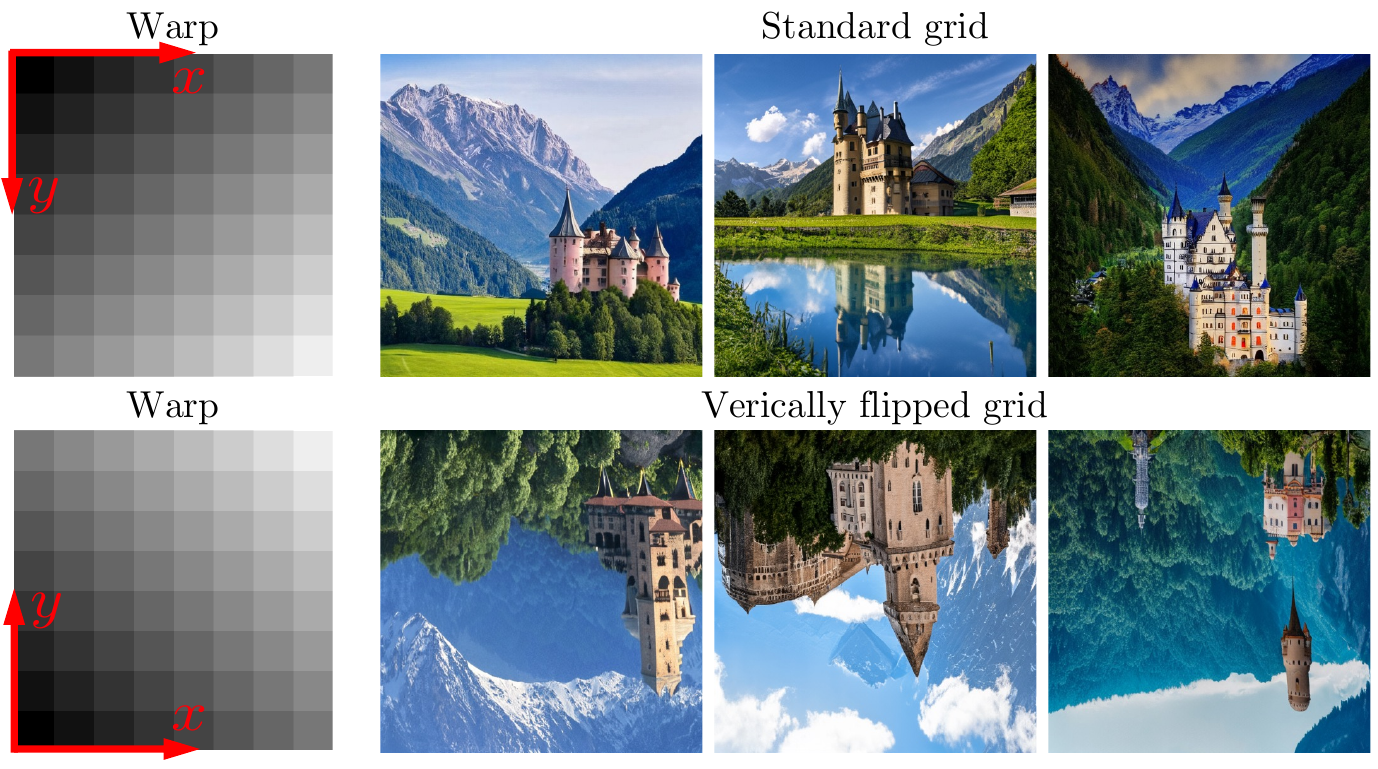}
    \caption{\textit{"Castle in the Alps"} generated with the standard grid (top), and with the upside-down vertical flipped axis (bottom).}
    \label{fig:vflip}
\end{figure}

While the model appears to be robust even for severe transforms not presented in the training warps family, it still suffers when the target deformation is too far from what is seen during the training. To illustrate the failure cases of our approach Figure \ref{fig:vflip} (bottom) shows the images generated with the coordinates grid flipped upside-down. While the model still roughly follows the global layout, it clearly hallucinates in some of the local regions, getting confused with following the "up" direction.

\vspace{-15pt}
\vspace{-25pt}

\section{Further Proofs}

As was noted in the main text, the proposed self-attention re-weighting has a clear interpretation for digit-value density. Just like in the main text, the \textit{spatial token} notation in the self-attention layer refers to the triplets $(k_i, q_i, v_i)$ of the query, key, and value at some spatial location $i$ in the intermediate self-attention layer. Here we show that for a digit-valued density $d \in \mathbb{N}$ at the spatial location $i$, the proposed re-weighting is equal to duplicating the correspondent key, query, and value $d$ times. Indeed, let us perform the straightforward computation. With no loss of generality, we assume $i = 1$.

The original sequence of self-attention layer inputs \\
$(k_1, q_1, v_1), \dots, (k_N, q_N, v_N)$ conducted the outputs $o_i = \sum\limits_{j=1}^{N} w_{ij} v_j$ with $w_{ij}$ being the weights computed as the softmax of the dot products $\left<q_i, k_j\right>$.

Once we duplicate the first spatial token, we add $d - 1$ copies of the triplets $(k_1, q_1, v_1)$ to the original spatial tokens set. This conducts the updated self-attention layer outputs $o_i' = \sum\limits_{j=1}^{N} w'_{ij} v_j$ where

$$
w'_{ij} = \frac{\exp(s_{ij})}{d \cdot \exp(s_{i1}) + \sum\limits_{l = 2}^N \exp(s_{il})}
$$
for $j \neq 1$, and 
$$
w'_{i1} = \frac{d \cdot \exp(s_{i1})}{d \cdot \exp(s_{i1}) + \sum\limits_{l = 2}^N \exp(s_{il})}
$$
Taking into account that $d \cdot \exp(s) = \exp(s + \ln d)$, this is equivalent to the equation 2 in the main text.

\section{Experiments Details}
Here we provide some implementation details. The training data distortion is implemented with the Open-CV python package \cite{opencv} distortion model. We set the distortion coefficients $k_1, k_2, p_1, p_2$ to be non-zero. The leading coefficient $k_1$ is sampled to be positive and negative with equal probabilities. All the coefficients are sampled from a uniform distribution with a random global scale $S$. For the first $80\%$ steps of training, we sample $S$ from the uniform distribution $U(0, 4)$, and on the last $20\%$ of steps we sample it from $U(4, 10)$.
For the positive $k_1$, all the coefficients are sampled from the distributions: $k_1 \sim U(0, 5),\ k_2 \sim U(0, 1.5), p_1, p_2 \sim U(0, 0.05)$ and then uniformly scaled by $S$. For the negative $k_1$ all the coefficients are sampled as: $k_1 \sim U(-0.1, -0.05),\ k_2 \sim U(-0.06, -0.01),\ p_1, p_2 \sim U(0, 0.00035)$ and also scaled by $S$. The focal center is randomly uniformly sampled from the box around the image center with the side $0.3$ multiplied by the image short side. With probability $0.3$ we don't apply any distortion. All these coefficients were selected manually to produce reasonable visual distortion and we have not performed a wider investigation of the best distortion coefficients distribution. While we perform $500 \times 10^3$ learning steps, we suppose that this number could be significantly reduced without much quality degradation. As our model modifies only the number of channels in the input convolutional layer of the base fine-tuned text-to-image model, it has the same performance as the base model.

Below we provide some base model details, though one should notice that the proposed method is model-agnostic. All experiments use models that are fine-tuned from a text-to-image model based on the LDM architecture \cite{latent_diffusion} with the image resolution $512\times 512$ and the latents shape $64 \times 64 \times 8$. This foundation model was trained on web-sourced image-text pairs filtered for text alignment and aesthetic quality from the WebLI \cite{webli} dataset. Some of the captions were automatically generated with an image captioning model.

\begin{figure}[h]
    \vspace{-10pt}
    \centering
    \includegraphics[width=0.91\textwidth]{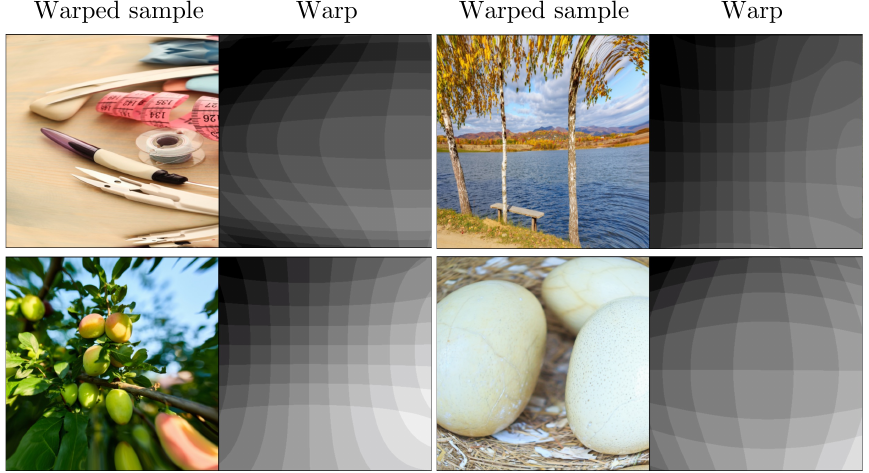}
    \caption{Training data samples for the per-pixel position conditioned model. Each warped image is equipped with the warping field illustrated to the right of it.}
    \label{fig:real}
    \condarxiv{\vspace{-25pt}}{\vspace{-25pt}}
\end{figure}

\begin{figure}
    \centering
    \includegraphics[width=0.91\textwidth]{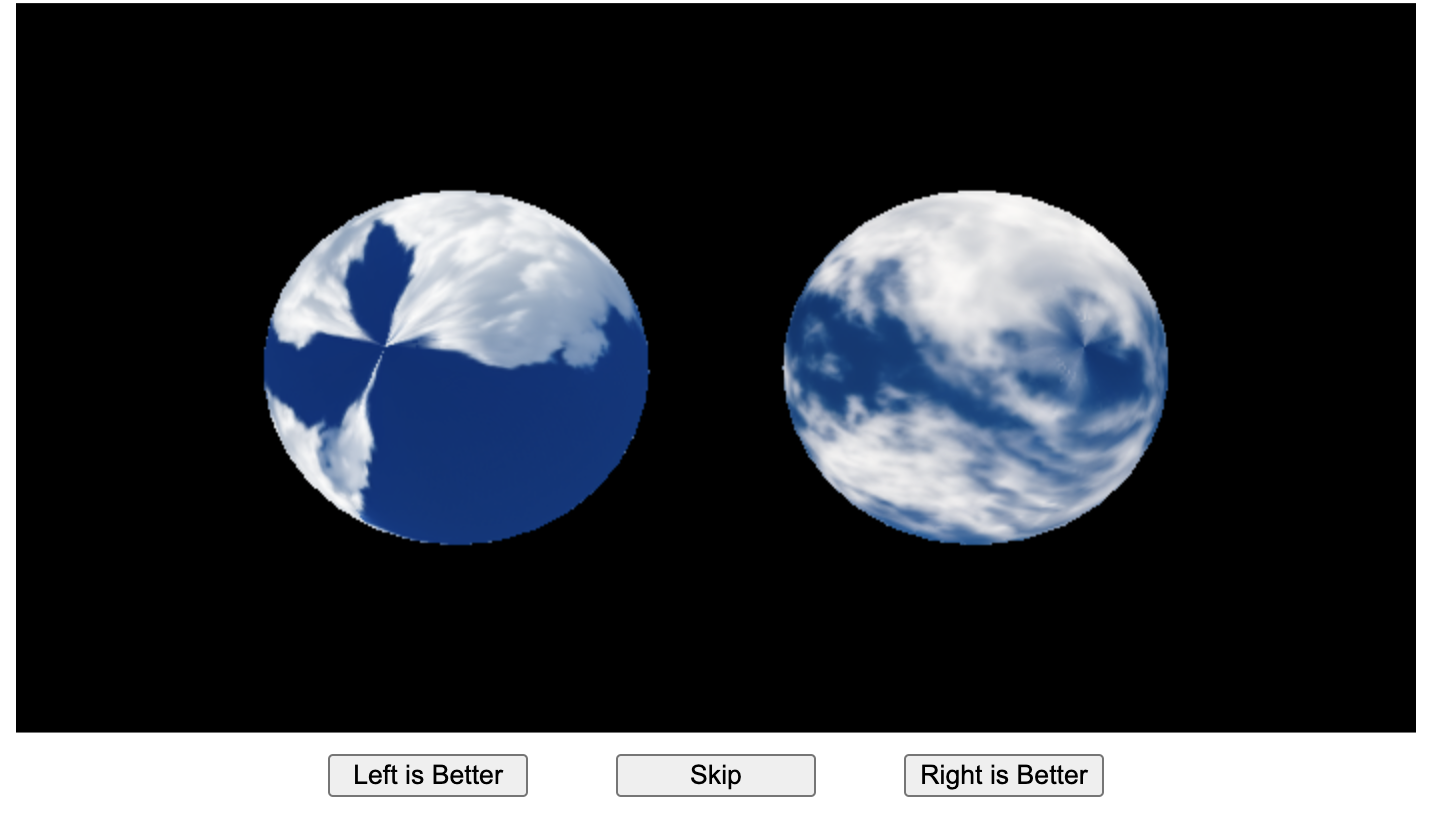}
    \caption{Screenshot of labeling interface for user study. Both spheres are rotatable.}
    \label{fig:label_interface}
\end{figure}

Figure \ref{fig:real} illustrates the training data samples consisted of a warped image and the normalized warped positions. Figure \ref{fig:label_interface} shows the labeling interface used for human evaluation.

\end{document}